\documentclass[times,twocolumn,final]{elsarticle}

\usepackage{media-arxiv}
\usepackage{framed,multirow}
\usepackage{adjustbox}
\usepackage[fleqn]{amsmath}
\usepackage{amssymb,bm}
\usepackage{latexsym}
\usepackage{soul}
\usepackage{ulem}
\usepackage{amsmath}
\usepackage{booktabs}
\usepackage{graphicx}
\usepackage{array}
\usepackage{makecell}
\usepackage{url}
\usepackage{xcolor}
\usepackage{hyperref}
\usepackage{vcell}
\usepackage{amssymb}
\usepackage{pifont}
\newcommand{\CheckmarkBold}{\ding{51}} 
\newcommand{\XSolidBrush}{\ding{55}} 
\definecolor{newcolor}{rgb}{.8,.349,.1}
\definecolor{highlight}{rgb}{0, 0, 0.0}
\journal{Medical Image Analysis}

\begin{document}

\verso{Zihao Zhao \textit{et~al.}}

\begin{frontmatter}

\title{CLIP in medical imaging: A survey}%

\author[1]{Zihao \snm{Zhao}\fnref{fn1}}
\author[1]{Yuxiao \snm{Liu}\fnref{fn1}}
\author[1]{Han \snm{Wu}\fnref{fn1}}
\fntext[fn1]{These authors contributed equally to this paper}
\author[1,2]{Mei \snm{Wang}}
\author[1]{Yonghao \snm{Li}}
\author[1,3]{Sheng \snm{Wang}}
\author[1]{Lin \snm{Teng}}
\author[1]{Disheng \snm{Liu}}
\author[1]{Zhiming \snm{Cui}\corref{cor1}}
\ead{cuizhm@shanghaitech.edu.cn}
\author[1]{Qian \snm{Wang}\corref{cor1}}
\ead{qianwang@shanghaitech.edu.cn}
\author[1,4,5]{Dinggang \snm{Shen}\corref{cor1}}
\ead{dgshen@shanghaitech.edu.cn}

\cortext[cor1]{Corresponding author: Zhiming Cui, Qian Wang, Dinggang Shen}
\address[1]{School of Biomedical Engineering \& State Key Laboratory of Advanced Medical Materials and Devices, ShanghaiTech University, Shanghai, China}
\address[2]{School of Biomedical Engineering, Southern Medical University, Guangzhou, China}
\address[3]{School of Biomedical Engineering, Shanghai Jiao Tong University, Shanghai, China}
\address[4]{Department of Research and Development, Shanghai United Imaging Intelligence Co., Ltd., Shanghai, China}
\address[5]{Shanghai Clinical Research and Trial Center, Shanghai, China}

\begin{abstract}
Contrastive Language-Image Pre-training (CLIP), a simple yet effective pre-training paradigm, successfully introduces text supervision to vision models. It has shown promising results across various tasks due to its generalizability and interpretability. The use of CLIP has recently gained increasing interest in the medical imaging domain, serving as a pre-training paradigm for image-text alignment, or a critical component in diverse clinical tasks. With the aim of facilitating a deeper understanding of this promising direction, this survey offers an in-depth exploration of the CLIP within the domain of medical imaging, regarding both refined CLIP pre-training and CLIP-driven applications. In this paper, we (1) first start with a brief introduction to the fundamentals of CLIP methodology; (2) then investigate the adaptation of CLIP pre-training in the medical imaging domain, focusing on how to optimize CLIP given characteristics of medical images and reports; (3) further explore practical utilization of CLIP pre-trained models in various tasks, including classification, dense prediction, and cross-modal tasks; and (4) finally discuss existing limitations of CLIP in the context of medical imaging, and propose forward-looking directions to address the demands of medical imaging domain. Studies featuring technical and practical value are both investigated. We expect this survey will provide researchers with a holistic understanding of the CLIP paradigm and its potential implications. The project page of this survey can also be found on \href{https://github.com/zhaozh10/Awesome-CLIP-in-Medical-Imaging}{Github}.
\end{abstract}


\end{frontmatter}



\section{Introduction}
\label{intro}
\begin{figure*}[!tbp]
    \centering
    \includegraphics[width=1.00\textwidth]{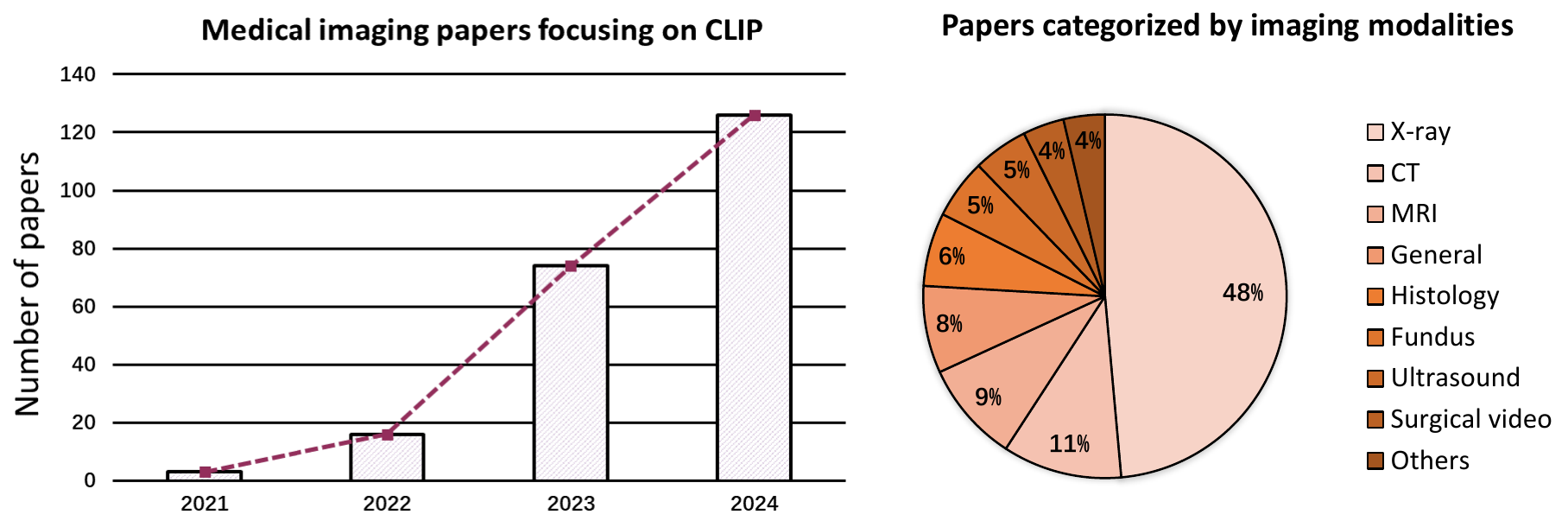}
    \caption{\textcolor{highlight}{\textbf{Left:} Rapid increase in medical imaging papers focusing on CLIP. \textbf{Right:}  
    Distribution of papers included in this survey, classified by different imaging modalities. Studies leveraging CLIP for developing generalist medical AI are classified as ``General''. Imaging modalities rarely used, including placenta, dermoscopy, and electrocardiograms, are grouped under "Others." }
    }
    \label{fig:growth_curve}
\end{figure*}

Despite substantial progress in vision intelligence over the last decade~\citep{he2016deep,tarvainen2017mean,dosovitskiy2020image,liu2021swin,liu2022convnet}, vision models were often trained only on vision-modality annotations and tasks~{\citep{ronneberger2015u,he2017mask,isensee2021nnunet,he2021masked}}.
{Although some of these methods have achieved human-level performance, their out-of-distribution performance is still far from satisfaction due to lack of alignment with human cognition~\citep{peterson2019human,battleday2020capturing, geirhos2020shortcut,peters2021capturing}.}
In contrast, the form of text supervision is naturally rich in semantics, and the corresponding language models, especially today's large language models~\citep{touvron2023llama2,xiong2023doctorglm,zhang2023huatuogpt}, typically contain a huge amount of human-level knowledge.
Hence, it is intuitive to integrate text supervision into vision tasks.

Drawing inspiration from contrastive pre-training, \citet{radford2021learning}  propose Contrastive Language Image Pre-training (CLIP), which learns interpretable visual representations from text supervision.
{Unlike most vision-only contrastive pre-training methods~\citep{chen2020simple,caron2021emerging,zbontar2021barlow},}
CLIP leverages both vision and language information.
Specifically, it considers the text caption as a linguistic view of the image and expects it to be inherently consistent with the image. Hence, it pulls paired image and text representations as close as possible in the latent space.
In this manner, the image-text pair is aligned through CLIP's vision and text encoders, and thus extensive knowledge has been encoded within the vision encoder.
Benefitting from image-text alignment, CLIP has learned extensive knowledge from text supervision and proven useful in a wide variety of areas, including image generation~\citep{vinker2022clipasso,ramesh2022hierarchical,yu2022towards,rombach2022high}, segmentation~\citep{li2022languagedriven,rao2022denseclip,luo2023segclip}, detection~\citep{bangalath2022bridging,lin2023gridclip}, and classification~\citep{zhou2022conditional,zhou2022learning,wang2023improving}. 
\begin{figure*}[!tbp]
    \centering
    \includegraphics[width=1.0\textwidth]{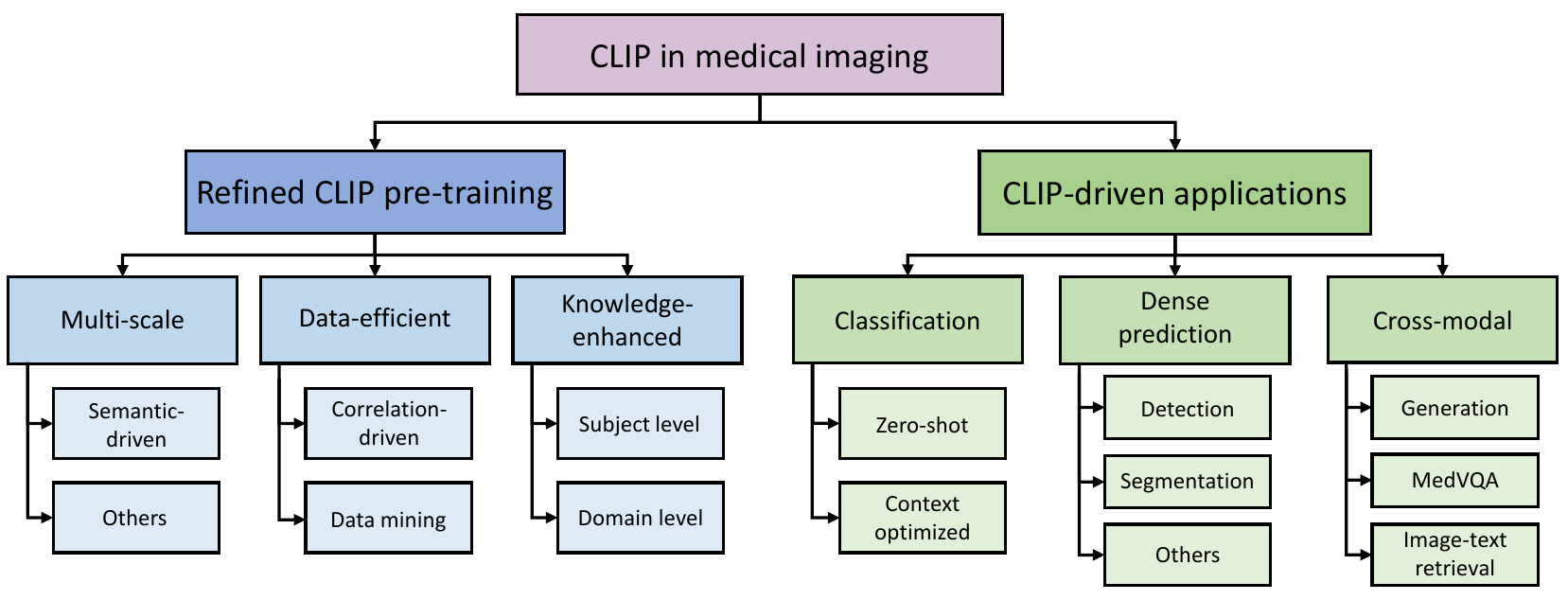}
    \caption{Taxonomy of studies focusing on CLIP in the field of medical imaging.}
    \label{fig:structure}
\end{figure*}

Recently, CLIP has also gained increasing attention in the field of medical imaging ~{\citep{eslami2021does,wang2022medclip,kim2024transparent}}.
\textcolor{highlight}{The left panel of} Fig.~\ref{fig:growth_curve} illustrates the growing trend of CLIP-related papers in the medical imaging domain, showing a several-fold increase from the second half year of 2021 to the second half of 2023, particularly with a booming trend in 2023. This can be attributed to its capability to align neural networks with human cognition, which meets the demands of
interpretability for AI in healthcare~\citep{lauritsen2020explainable,tjoa2020survey,XIE2021101985}.
{Although previous studies have attempted to improve interpretability via additional expert-level annotations, such as bounding boxes~\citep{luo2022rethinking,ouyang2020learning,tanida2023interactive,muller2023anatomy} and segmentation masks~\citep{mehta2018net,zhou2019collaborative}, collecting these annotations is laborious
and time-consuming~\citep{rahimi2021addressing,wang2022follow,qu2023abdomenatlas}, which makes them challenging to scale.}
~{On the contrary, text reports, routinely produced and collected in clinical
practice by medical professionals in clinical practice~\citep{huang2021gloria,wang2022multi,xie2023medim} offer valuable expert-level knowledge with few additional costs, which positions CLIP as a more promising solution.}

\textbf{Motivation.}
Although many studies have been devoted to customizing CLIP for medical image analysis, this customization raised new challenges that remain unsolved. To encourage and facilitate CLIP-based research in medical imaging domain, a survey of the existing literature would be beneficial.  Thus, we extensively review more than 200 existing methods in the field for providing an insightful survey.

\textcolor{highlight}{
\textbf{Search strategy.}
We initiated our search across academic platforms, including Google Scholar, DBLP, ArXiv, and IEEE Xplore. Utilizing keywords such as ``CLIP'', ``image-text alignment'', and ``medical imaging'', our initial exploration covers a diverse range of sources, such as high-impact journal papers, conference/workshop papers, and preprints still under review. 
We excluded studies using CLIP solely as a baseline method and studies not focusing on medical imaging. Given that the CLIP text encoder has become a key component in diffusion models~\citep{kazerouni2023diffusion} following the success of Stable Diffusion~\citep{rombach2022high}, we also excluded papers that focused primarily on diffusion models rather than CLIP.
In addition, imaging modalities such as surgical videos and ultrasound are often presented in the form of video clips. As a result, some studies focused on these modalities were mistakenly retrieved, even though not using CLIP-related techniques. These false-retrieved studies were manually reviewed and excluded, while relevant studies that did adopt CLIP were retained.
After applying these selection criteria, we identified a total of 224 papers for this survey. The distribution of selected papers, classified by their respectively-used imaging modalities, is shown on the right panel of Fig.~\ref{fig:growth_curve}, with X-ray-based studies occupying the majority.
}

\textbf{{Taxonomy.}}
To streamline our discussion on CLIP-related work in medical imaging domain, we taxonomize them into two categories in this survey, i.e., (1) refined CLIP pre-training and (2) CLIP-driven applications, as illustrated in Fig.~\ref{fig:structure}.
This first category of studies focuses on adapting the pre-training paradigm of CLIP from web-crawled image-caption pairs to medical images and their respective reports. The second category of studies tends to directly adopt pre-trained CLIP models to improve the interpretability and robustness of deep learning models in various clinical tasks e.g., thoracic disease diagnosis and multi-organ segmentation~\citep{tiu2022expert,pellegrini2023xplainer,liu2023clip}.

 \textbf{Relevant surveys.}
 There are some concurrent surveys sharing a similar scope as ours. 

 For example, a survey paper by~\citep{shrestha2023medical} focuses on medical vision language pre-training. 
 The differences between their survey and our survey are:
1) They mainly explore various vision language pre-training architectures, including masked prediction~\citep{khare2021mmbert}, contrastive~\citep{huang2021gloria}, matching prediction~\citep{moon2022multi}, and hybrid architectures~\citep{wang2021self}, with less emphasis on CLIP-style contrastive pre-training; 2) Their primary focus is on large-scale pre-training, lacking discussions on 
clinical tasks. Due to scarcity of large public medical image-text datasets, the impact of their survey may be limited. 
In contrast, our survey delves into CLIP-driven applications, which are more feasible in real-world settings due to lower data requirements and more values for real-world clinical tasks.
Additionally, ~\cite{azad2023foundational} explore foundational models in medical imaging, with CLIP also included in their survey paper. The main difference from our survey paper is that they mainly cover vision-only, language-only, and vision-language foundational models, whereas we focus exclusively on CLIP, a vision-language foundational model. Furthermore, their survey also lacks discussions on clinical applications, for which we cover extensively. Accordingly, our survey (compared to the above two survey papers) will provide a more insightful and deeper survey, in terms of both techniques and clinical applications.

\textbf{Contribution.} 
{In summary, our contributions are as follows:}
\begin{itemize}
    \item
    To the best of our knowledge, this paper is the first review of CLIP in medical imaging, aiming to provide a timely summary and insight for potential studies in this rapidly evolving area.
    \item We provide thorough coverage of existing studies and also a multi-level taxonomy to serve different needs of readers.
    \item Furthermore, we discuss the issues and open questions in this field. We also pinpoint new trends and propose future directions for further exploration.
\end{itemize}

\textbf{Paper organization.}
The rest of the paper is organized as follows. Section~\ref{sec: background} provides preliminary knowledge of CLIP and its variants. 
In Section~\ref{sec: pretrain}, we present a systematic analysis of how to adapt CLIP to the medical imaging field, from the perspectives of key challenges and corresponding solutions. Section~\ref{sec: downstream} covers various clinical applications of pre-trained CLIP and compares CLIP-driven methods with early methods/solutions. 
Section~\ref{sec: discussion} further discusses existing limitations, as well as potential research directions.
We finally conclude this paper in Section~\ref{sec: conclusion}.

\section{Background}\label{sec: background}
CLIP-related research has advanced rapidly in recent years.
In this section, we provide a brief overview of CLIP, as well as its generalizability and multiple variants.
Additionally, we summarize datasets of medical image-text pairs that are publicly available and usable to CLIP.

\subsection{Contrastive Language-Image Pre-training}\label{clip_theory}
 CLIP (\textbf{C}ontrastive \textbf{L}anguage-\textbf{I}mage \textbf{P}re-training) is a pre-training method developed by OpenAI, designed to bridge the gap between images and texts. It jointly optimizes a vision encoder and a text encoder, ensuring that image-text pairs are closely aligned in a shared latent space. 
 Unlike other methods that rely on extensive manual supervision or complex structures, CLIP follows the principle of Occam's Razor~\citep{blumer1987occam}, favoring simplicity for effective results.

\begin{figure}[tbp]
    \centering
    \includegraphics[width=0.5\textwidth]{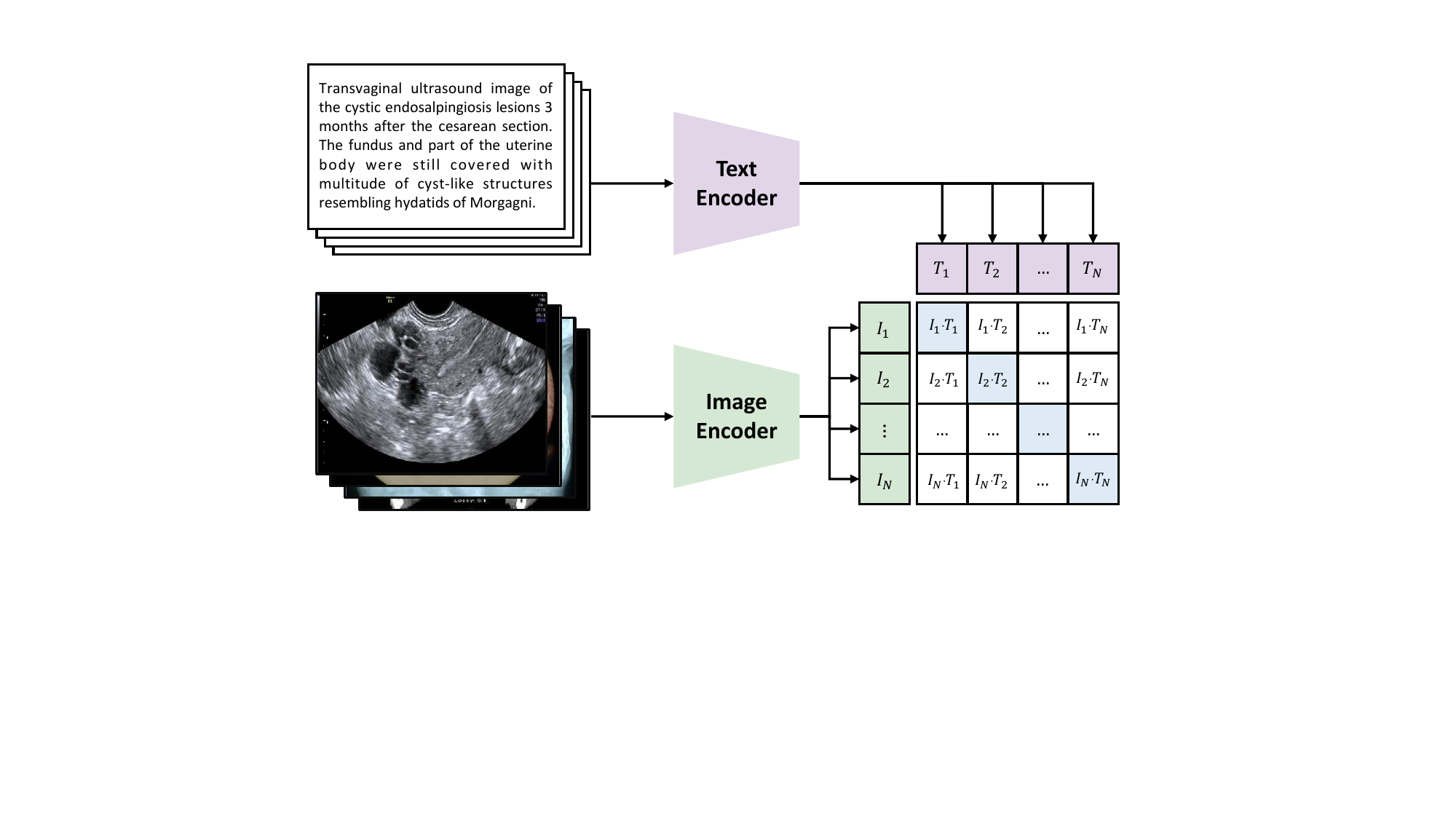}
    \caption{Illustration of CLIP in medical imaging, with an example from the PMC-OA dataset.}
    \label{fig:clip-train}
\end{figure}

\textbf{Architecture.} 
In terms of its architecture, CLIP seamlessly integrates a vision model with a language model. The visual component can be based on either ResNet~\citep{he2016deep} or Vision Transformer (ViT)~\citep{dosovitskiy2020image}, while the language encoder is rooted in a transformer-based model like BERT~\citep{kenton2019bert}. As illustrated in Fig.~\ref{fig:clip-train}, it receives a batch of images and their corresponding text descriptions as input in each iteration.
Following the encoding process, the embeddings are normalized and mapped to a joint image-text latent space. That is, the input images and texts are encoded into $\mathit{I} \in \mathbb{R}^{N \times D}$ and $\mathit{T} \in \mathbb{R}^{N \times D}$, respectively, where $N$ denotes batch size and $D$ represents embedding dimensionality.

\textbf{Contrastive pre-training.}
In CLIP, contrastive pre-training plays a crucial role in aligning image-text pairs. Diverging from conventional models that are sculpted for a singular and pre-defined task, CLIP's learning trajectory revolves around contrastive pre-training between paired image-text information.
In particular, $N^2$ image-text pairs can be constructed given a batch size of $N$, among which there are $N$ matched image-text pairs (positive pairs, as highlighted in blue in Fig.~\ref{fig:clip-train}) and $(N^2-N)$ unmatched image-text pairs (negative pairs). The pre-training objective for the image encoder is hence denoted as
\begin{align}
\mathcal{L}_\textrm{img} = -\frac{1}{N} \sum_{i=1}^{N}\log  \frac{\exp(\Phi(I_i, T_i)/\tau)}{\sum_{j=1}^{N} \exp(\Phi(I_i, T_j)/\tau)},
\label{eq:loss_img}
\end{align}
where $\Phi(\cdot,\cdot)$ indicates cosine similarity, $\tau$ is a learnable temperature parameter, $I_i$ and $T_i$ represent the $i_\textrm{th}$ image embedding and text embedding, respectively. The objective for the text encoder is defined symmetrically:
\begin{align}
\mathcal{L}_\textrm{txt} = -\frac{1}{N} \sum_{i=1}^{N}\log  \frac{\exp(\Phi(T_i, I_i)/\tau)}{\sum_{j=1}^{N} \exp(\Phi(T_i, I_j)/\tau)}.
\label{eq:loss_txt}
\end{align}
The total optimization objective of CLIP is hence calculated via the average of~\eqref{eq:loss_img} and~\eqref{eq:loss_txt}:
\begin{align}
    \mathcal{L}_\textrm{total}=\frac{\mathcal{L}_\textrm{img}+\mathcal{L}_\textrm{txt}}{2}.
\end{align}

\begin{figure}[tbp]
    \centering
    \includegraphics[width=0.5\textwidth]{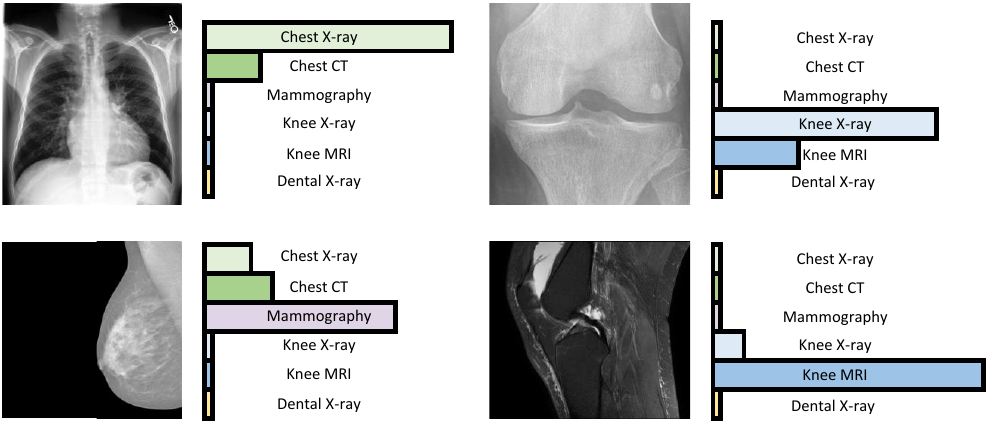}
    \caption{Illustration of CLIP's generalizability via domain identification.}
    \label{fig:clip-explainable}
\end{figure}
\textbf{Zero-shot capability and generalizability.}
Since CLIP is pre-trained to predict whether an image matches a textual description, it naturally lends itself to zero-shot recognition. 
This process is accomplished by comparing image embeddings with text embeddings, which correspond to textual descriptions specifying certain classes of interest. 
Let $I$ represent the image features extracted by the image encoder for a given image \(x\), and let \(\{ W_i \}_{i=1}^K\) be the set of class embeddings generated by the text encoder. Here, \(K\) denotes the number of classes, and each \(W_i\) is derived from a text prompt resembling ``a photo of a [CLASS]", where the class token is substituted with the specific class name. 
The probability of prediction is then calculated as follows:
\begin{equation} \label{eq:prob}
p(y=i|I) = \frac{\exp(\Phi(I, W_i)/\tau)}{\sum_{j=1}^K \exp(\Phi(I, W_j)/\tau)},
\end{equation}
where \(\tau\) is a temperature parameter learned during pre-training, and \(\Phi(\cdot, \cdot)\) represents the cosine similarity.
In contrast with traditional classifier learning methods where closed-set visual concepts are learned from scratch, CLIP pre-training allows for the exploration of open-set visual concepts through the text encoder. This leads to a broader semantic space and, consequently, makes the learned representations more transferable to downstream tasks.

The generalizability of the CLIP pre-trained model becomes evident when applied to specialized areas such as medical imaging. Although originally trained on internet images and their textual captions, CLIP has demonstrated the capability to recognize and categorize medical images. Fig. \ref{fig:clip-explainable} illustrates the generalizability of CLIP via domain identification, where the class token in the text prompt is substituted with the specific class name, such as ``Chest X-ray", ``Mammography", ``Knee X-ray",  or ``Dental X-ray".
Its zero-shot inference capability allows it to identify the domain of a given medical image without explicit prior training on such datasets. 
While further studies and validations are necessary, the preliminary findings suggest that the zero-shot capability of a pre-trained CLIP could reduce the dependency on extensive labeled medical datasets and pave the way for more efficient and generalizable AI-driven diagnostic tools.
\begin{table*}[tbp]\footnotesize
    \centering
    \caption{Summary of publicly available medical image-text datasets.}
    \label{tab:dataset}
    \begin{tabular}{lcllccc}
    \toprule
        Dataset & Domain & Image&Text & Source & Language & Pre-trained CLIP\\
    \midrule
        ROCO~\citep{pelka2018radiology} & \textcolor{highlight}{General} & 87K& 87K& Research papers & English& \href{https://huggingface.co/flaviagiammarino/pubmed-clip-vit-base-patch32}{PubMedCLIP} \\
        MedICaT~\citep{subramanian2020medicat} & \textcolor{highlight}{General} & 217K&  217K& Research papers&English& /\\
        PMC-OA~\citep{lin2023pmc} & \textcolor{highlight}{General} & 1.6M &1.6M & Research papers&English& \href{https://huggingface.co/ryanyip7777/pmc_vit_l_14}{PMC-CLIP}\\
        ChiMed-VL~\citep{liu2023qilin} & \textcolor{highlight}{General} & 580K &580K & Research papers&English \& Chinese& /\\
        FFA-IR~\citep{li2021ffa} & Fundus & 1M&10K& Medical reports & English \& Chinese&/ \\
        \textcolor{highlight}{FUNDUS}~\citep{silva2025foundation} & \textcolor{highlight}{Fundus} & \textcolor{highlight}{288K}&\textcolor{highlight}{288K}& \textcolor{highlight}{Synthesized text descriptions} & \textcolor{highlight}{English}&~\href{https://github.com/jusiro/FLAIR}{FLAIR} \\
        \textcolor{highlight}{MM-Retinal}~\citep{wu2024mm} & \textcolor{highlight}{Fundus} & \textcolor{highlight}{4.3K}&\textcolor{highlight}{4.3K}& \textcolor{highlight}{Textbooks} & \textcolor{highlight}{English \& Chinese}&~\href{https://github.com/lxirich/MM-Retinal}{KeepFIT} \\
        PadChest~\citep{bustos2020padchest} & Chest X-ray & 160K & 109K& Medical reports& Spanish&/ \\
        MIMIC-CXR~\citep{johnson2019mimic} & Chest X-ray & 377K & 227K& Medical reports& English& \href{https://huggingface.co/microsoft/BiomedVLP-CXR-BERT-specialized}{BioViL}\&~\href{https://huggingface.co/microsoft/BiomedVLP-BioViL-T}{BioViL-T}\\
        Chexpert-Plus~\citep{chambon2024chexpert} & Chest X-ray & 223K & 187K& Medical reports& English& \href{https://huggingface.co/StanfordAIMI/XrayCLIP__vit-b-16__laion2b-s34b-b88k}{XrayCLIP}\\
        CT-RATE~\citep{hamamci2024foundation} & Chest CT & 50K & 50K & Medical reports& English & ~\href{https://huggingface.co/datasets/ibrahimhamamci/CT-RATE}{CT-CLIP}\\
        INSPECT~\citep{huang2023inspect} & Chest CT & 23K & 23K & Medical reports& English & /\\
        \textcolor{highlight}{AbdomenAtlas 3.0}~\citep{bassi2025radgpt} & \textcolor{highlight}{Abdominal CT} & \textcolor{highlight}{9K} & \textcolor{highlight}{9K} & \textcolor{highlight}{Medical reports \& Synthetic data}& \textcolor{highlight}{English} & /\\
        OpenPath~\citep{huang2023visual} & Histology & 208K& 208K & Social media& English& \href{https://huggingface.co/vinid/plip}{PLIP}\\
        Quilt-1M~\citep{ikezogwo2023quilt} & Histology & 1M & 1M & Research papers \& Social media&English& \href{https://huggingface.co/wisdomik/QuiltNet-B-32}{QuiltNet}\\
        \textcolor{highlight}{PathGen-1.6M}~\citep{sun2025pathgenm} & \textcolor{highlight}{Histology} & \textcolor{highlight}{1.6M} & \textcolor{highlight}{1.6M} & \textcolor{highlight}{Medical reports \& Synthetic data}&\textcolor{highlight}{English}&  \href{https://pan.quark.cn/s/62fe3dc65291#/list/share}{PathGen-CLIP}\\
    \bottomrule
    \end{tabular}
\end{table*}

\subsection{Variants of CLIP}
After providing a concise overview of CLIP in the above, we hereby introduce several variants of CLIP with practical applications in the area of medical imaging, which takes a step further by \textit{not only} recognizing items in images \textit{but also} understanding their specific details and descriptions. 

Following the philosophy of CLIP, GLIP~\citep{li2022grounded} reformulates detection as a grounding task by aligning each region or bounding box with corresponding text phrases. It simultaneously trains both an image encoder and a language encoder to accurately predict the associations between regions and words. A fusion module is further proposed to enhance the alignment between image and text information, improving the model's ability to learn a language-aware visual representation.
Pre-trained specifically at the object level, GLIP has demonstrated remarkable performance, even comparable to fully supervised methods in zero-shot object detection and phrase grounding tasks. 

Meanwhile, CLIPSeg~\citep{luddecke2022image} and CRIS~\citep{wang2022cris} extend CLIP to the area of segmentation. 
CLIPSeg has fixed the pre-trained CLIP image encoder and text encoder while introducing a trainable decoder for the segmentation task. 
The encoded image and text prompt are fused, and then input into the trainable decoder to generate the predicted segmentation mask. 
Conceptually aligned, a similar paradigm is proposed in CRIS. These representative variants have been favored by generalizability, further showcasing the adaptability of CLIP.

The potential of these variants lies in their attention to content details. They could combine visual and textual information to provide a more nuanced understanding of medical images.
For medical imaging, it is critical to identify subtle features, e.g., related to tumors or bone fractures.
Such techniques could offer significant benefits, potentially being able to spatially locate clinical findings given the provided prompt, like ``malignant mass" or ``calcification."

\subsection{Medical image-text dataset}
{Large-scale datasets are essential to ensure the quality of pre-training in medical imaging~\citep{azizi2021big,cherti2023reproducible,huang2023stunet,wang2023foundation}.}
Therefore, we summarize those publicly available medical datasets in Table~\ref{tab:dataset}. These datasets encompass various medical domains and data sources. For each of them, we also signify whether there is a publicly available CLIP model pre-trained on this dataset. 
ROCO~\citep{pelka2018radiology}, MedICaT~\citep{subramanian2020medicat}, PMC-OA~\citep{lin2023pmc}, and ChiMed-VL~\citep{liu2023qilin} are the four large-scale datasets sourced from research papers. They collect and filter biomedical figure-caption pairs from open-access research papers via PubMed Central\footnote{\href{https://www.ncbi.nlm.nih.gov/pmc/tools/ftp/}{https://www.ncbi.nlm.nih.gov/pmc/tools/ftp/}}. Since research papers could cover a wide range of topics, the resulting datasets are composed of diverse medical images, including X-ray, PET, MRI, etc.
FFA-IR~\citep{li2021ffa}, PadChest~\citep{bustos2020padchest}, and MIMIC-CXR~\citep{johnson2019mimic} are collected from daily medical reports.
In clinical practice, a diagnostic report is often derived from multiple images. 
Therefore, the number of image samples and text samples shows a significant disparity, especially in the case of the FFA-IR dataset.
OpenPath~\citep{huang2023visual} obtains histology images and their paired captions from Twitter, a social media platform. The proposed histology foundation model, PLIP, has shown impressive performance, illuminating significance of social media-derived data. 
Following OpenPath, Quilt-1M~\citep{ikezogwo2023quilt} extracts image-text pairs from both research papers and social media platforms like YouTube.
\textcolor{highlight}{
Differently, PathGen-1.6M~\citep{sun2025pathgenm} expands the scale of image-text pairs by leveraging a pre-trained image captioning model, making it the largest image-text dataset for histological images as of February 2025.~\cite{silva2025foundation} adopt a similar strategy and craft the FUNDUS dataset by generating text descriptions from categorical labels.
AbdomenAtlas 3.0~\citep{bassi2025radgpt} stands out as the first publicly available dataset that offers high-quality abdominal CTs alongside paired radiology reports. It additionally provides voxel-level segmentation annotations. This innovative dataset is anticipated to inspire further research in the literature.
}
Note that not all of these datasets are in English. PadChest is provided in Spanish, while FFA-IR, ChiMed-VL, and \textcolor{highlight}{MM-Retinal} have their respective Chinese versions. 
{The discrepancy between different languages may introduce language bias~\citep{zhou2021uc2} when pre-training on a combination of datasets from various language communities, which has recently motivated the study on cross-lingual pre-training in medical imaging community~\citep{wan2023medunic}.}

\section{CLIP in medical image-text pre-training}\label{sec: pretrain}
{Most of CLIP models are trained on web-crawled data~\citep{radford2021learning,cherti2023reproducible,EVA-CLIP}, without much consideration on special characteristics of medical imaging and reports.}
{To satisfy the needs of specialized foundational models~\citep{zhang2023challenges}, several efforts have been made to adapt the paradigm of CLIP to a specific medical imaging domain, e.g., Chest X-ray, Brain MRI, etc.}

In this section, we describe specific challenges of medical image-text pre-training, and provide a taxonomy of existing studies given their solutions. 
Representative methods included in this section are shown in Table~\ref{tab:pre-training-paper}. Their pre-trained imaging domains, specific taxonomy, evaluation tasks, and noteworthy issues are demonstrated. Evaluation tasks here mean that the quality of pre-trained vision model and text model are evaluated by directly observing their performance on targeting tasks without much modification, which is different from CLIP-driven applications to be mentioned in Section~\ref{sec: downstream}.
\subsection{Challenges of CLIP pre-training}
\begin{figure}
    \centering
    \includegraphics[width=\columnwidth]{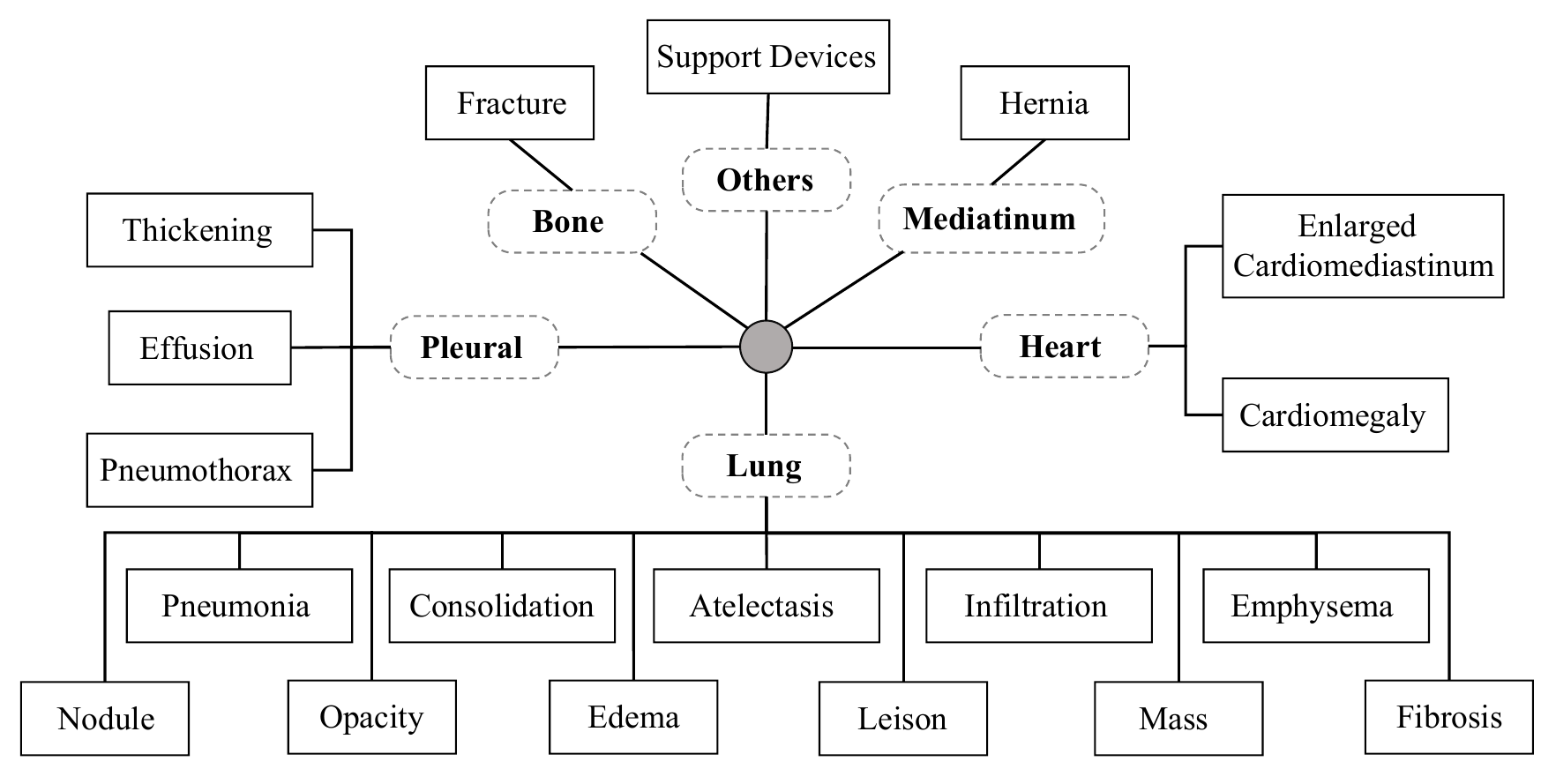}   
    \caption{An illustration of hierarchical dependencies among clinical findings in chest X-rays (sourced from~\citet{huang2023kiut}). Solid boxes indicate clinical findings while dotted boxes represent organs or tissues.}
    \label{fig:cxr_graph}
\end{figure}

CLIP is initially proposed on natural image datasets, which may lead to suboptimal performance 
on medical imaging due to three key challenges.
\begin{figure}[tbp]
    \centering
    \includegraphics[width=0.45\textwidth]{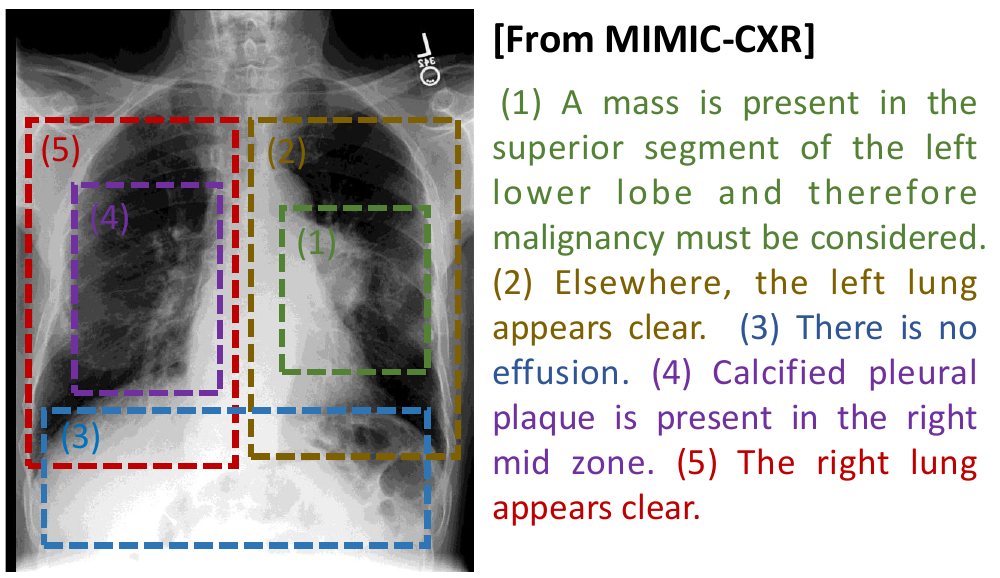}
    \caption{Demonstration of multi-scale features of medical image-text pairs. The medical report is composed of several sentences, with each sentence focusing on region-level features instead of global-level features. Sentences are independent of each other, and hold different levels of significance. }
    \label{fig:fine_grained}
\end{figure}
\begin{itemize}
    \item \textbf{Multi-scale features.}
    One main difference between medical images and natural images is the significance of multi-scale features. 
    Besides global-level vision features, local-level vision features also matter in the interpretation of medical images{~\citep{huang2021gloria,lv2023local,chaitanya2020contrastive,dubey2015local}. For example, some abnormalities or lesions such as lung nodule may only occupy small proportions in a chest radiograph, yet their crucial visual cues can significantly influence diagnostic results~\citep{austin1992missed,lo1995artificial,samei2003subtle,li2020multi}.}
    Besides image information, the corresponding text information is also characterized by multi-scale text features~{\citep{morid2016classification,luo2020identifying,yang2021writing,tanida2023interactive}}. 
    Medical reports tend to be more complex than natural image captions~{\citep{chen2020generating,miura2021improving}}. For example, natural image captions are typically concise and provide an overview of global features of the image. In contrast, as depicted in Fig.~\ref{fig:fine_grained}, medical reports consist of multiple sentences, with each sentence describing image findings in a specific region. For instance, the first sentence (highlighted in green) in Fig.~\ref{fig:fine_grained} describes the presence of a mass, which is essential for accurate diagnosis.
    Generally speaking, besides global-level image-text contrast, both local-level image features and local-level text features should be taken into consideration during the pre-training, posing challenges to the baseline CLIP pre-training, where image and text information are aligned solely at the global level.
    \item \textbf{Data scarcity.} Unlike natural image-text datasets, which can easily reach billion-scale~\citep{schuhmann2022laion,zhai2022scaling,zhu2023multimodal}, medical datasets with paired images and reports~\citep{johnson2019mimic,ikezogwo2023quilt} hold a relatively limited scale. 
    As the scale of datasets can have a significant impact on CLIP-style pre-training according to scaling laws~\citep{cherti2023reproducible}, limited medical data can hinder its performance in medical imaging domain.
    \item \textbf{High demands for specialized knowledge.} 
     The hierarchical dependencies among various clinical concepts can be intricate and highly specialized{~\citep{zhang2020radiology,jain2021radgraph}}. 
     As depicted in Fig.~\ref{fig:cxr_graph}, the graph is constructed based on the expert viewpoint of chest X-rays, considering correlations, characteristics, and occurrence locations of clinical findings~\citep{huang2023kiut}.
     Lack of in-depth understanding of medical concepts may lead to degraded performance when facing data from shifted distributions, or even shortcut solutions~\citep{geirhos2020shortcut}.
    Hence, in order to improve the reliability and robustness, explicitly incorporating knowledge during the process of pre-training may provide a viable solution.    
\end{itemize}

\begin{figure*}
    \centering
    \includegraphics[width=1.0\textwidth]{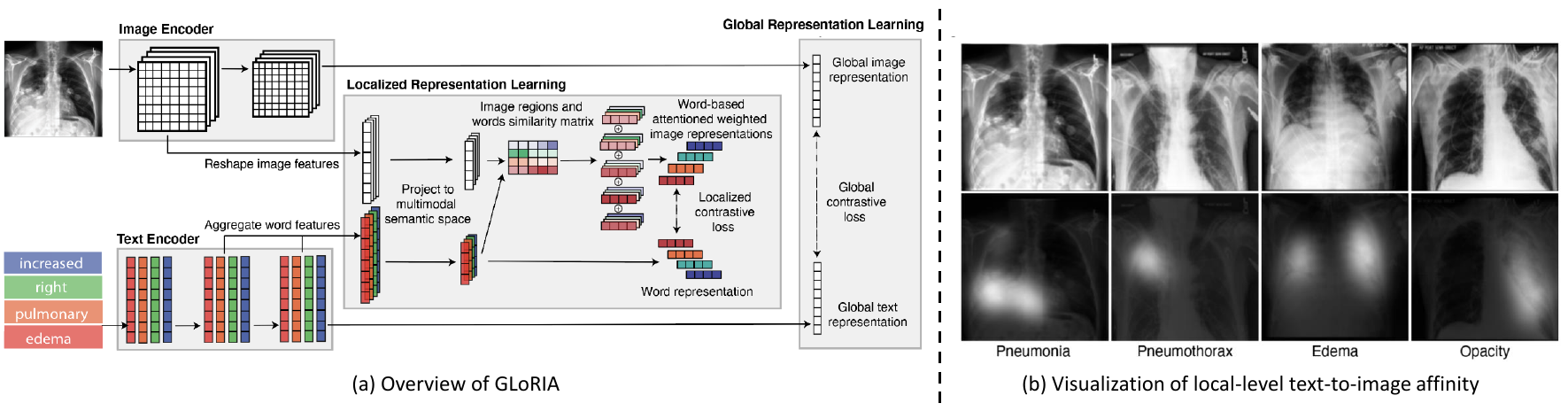}
    \caption{Illustration of the semantic-driven contrast proposed by GLoRIA~\citep{huang2021gloria}. (a) Overview of GLoRIA, which performs multi-scale image-text alignment based on cross-modal semantic affinity. 
     (b) Visualization of the semantic affinity learned by GLoRIA.} 
    \label{fig:gloria}
\end{figure*}
These challenges highlight the impracticality of directly applying CLIP pre-training on medical image-text datasets, motivating related work to improve CLIP-style pre-training in the medical imaging domain.

\subsection{Multi-scale contrast}\label{sec: global-local}
Although some early-stage studies~\citep{zhang2022contrastive,zhou2022generalized} attempt to extend CLIP pre-training to the medical imaging domain, they still follow the global-level contrast proposed in~\citet{radford2021learning} and hence show sub-optimal performance on tasks such as semantic segmentation and object detection.
To address the issue, several studies have tried to perform multi-scale contrast in pre-training. 
\begin{figure}[tbp]
    \centering
    \includegraphics[width=0.45\textwidth]{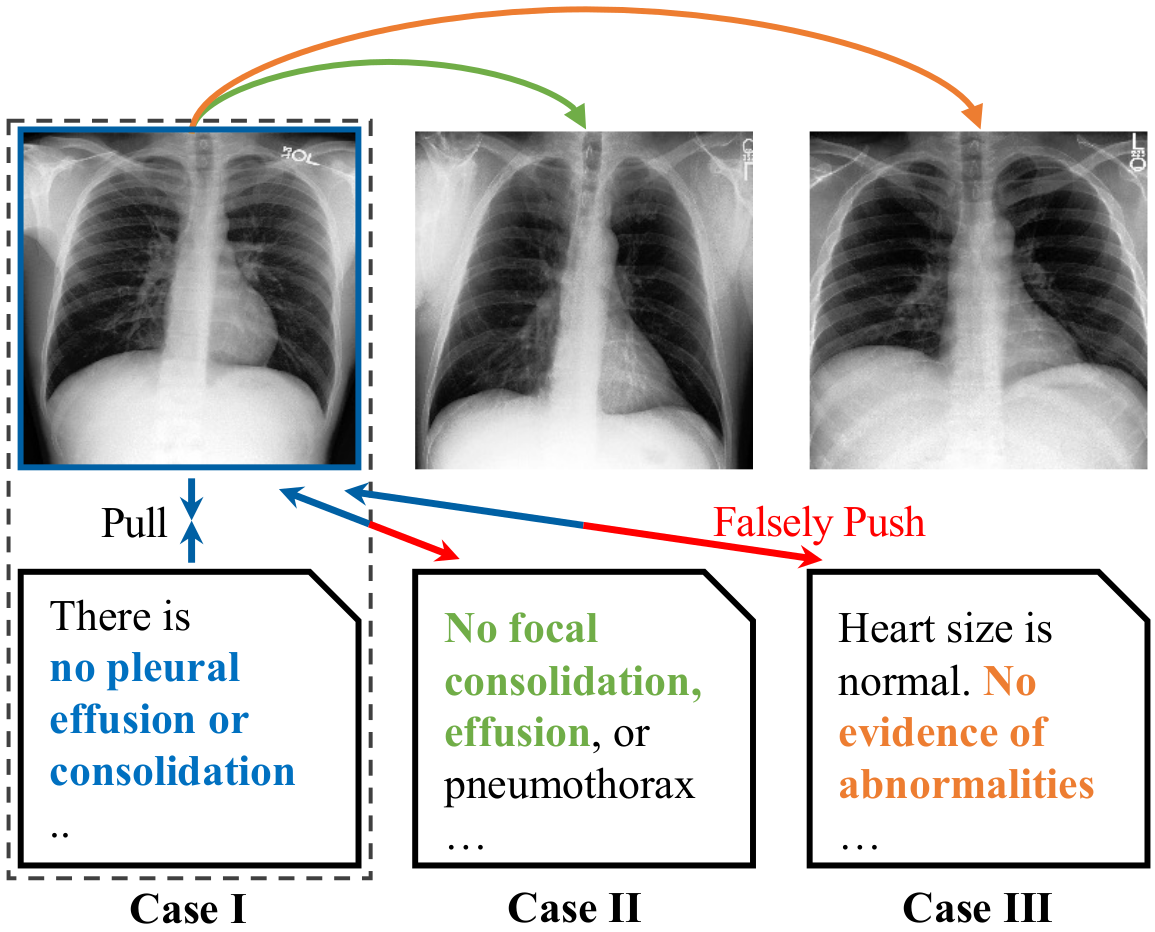}
    \caption{Illustration of false-negative pairs~\citep{liu2023improving}. In CLIP pre-training, a positive pair is defined as an image and its corresponding report, whereas all other reports are considered negatives.  This can result in false negatives, where semantically similar reports from different subjects are mistakenly considered as negative pairs.}
    \label{fig:false_negative}
\end{figure}

\textbf{Semantic-driven contrast.}~\cite{huang2021gloria} made a pioneering contribution in this domain by introducing the concept of semantic-driven multi-scale contrast. 
The proposed GLoRIA follows a similar paradigm as CLIP to implement global-level contrast, yet it distinguishes itself by implementing local-level contrast between each word representation and its semantically similar visual counterparts.
As illustrated in Fig.~\ref{fig:gloria}(a), GLoRIA defines each word and each image sub-region as local text and image features, respectively.
To perform local-level contrast, it calculates semantic similarity between word-wise text features and sub-region-wise image features. After obtaining semantic similarity, GLoRIA leverages it to compute weighted summation over all local image features and hence gets an attention-weighted local image representation for each word representation, where ``attention'' used in their paper denotes semantic similarity. The word representation and the weighted image representation are semantically similar and hence are pulled closer in the latent space via localized contrastive loss.
Since local-level contrast is implemented at the word level, GLoRIA would accumulate all localized contrastive loss as the total local-level contrast objective for medical report.
Fig.~\ref{fig:gloria}(b) demonstrates the effectiveness of semantic affinity learned by GLoRIA. The text-to-image semantic affinity, visualized as a heatmap, is able to correctly identify related image sub-regions for a given word. For instance, the semantic affinity based on the word ``Pneumonia” correctly localizes regions of the right lower lobe containing heterogenous consolidative opacities indicative of pneumonia. Additionally, attention weights associated with ``Pneumothorax" accurately emphasize lucency in the right lung apex, indicative of pneumothorax. Analogous results can be also observed for ``Edema" and ``Opacity", highlighting the effectiveness of local-level alignment.
\begin{figure*}[tbp]
    \centering
    \includegraphics[width=1.0\textwidth]{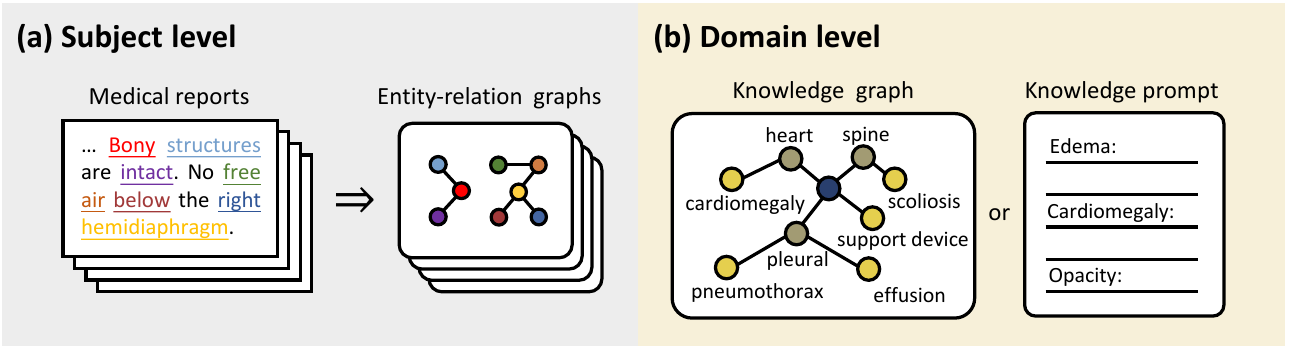}
    \caption{Illustration of knowledge enhancement at different levels (with the chest X-ray as an example). (a) At the subject level, external knowledge aids in converting medical reports into entity-relation graphs, which elucidate the causal relationships among medical entities in the report. (b) At the domain level, the domain knowledge is presented as a knowledge graph or descriptive knowledge prompt, directly providing human prior guidance for pre-training.}
    \label{fig:knowledge-enhanced}
\end{figure*}

The semantic-driven multi-scale contrast proposed by~\citet{huang2021gloria} is intuitive, but it still has some apparent weaknesses.
(1) The local-level contrast is designed asymmetrically. Its contrastive objective is optimized between text and attention-weighted sub-region features. This only guarantees alignment from text to image, dismissing alignment from image to text. 
(2) While computing text-to-image local features through weighted summation of local image features is intuitive, it may struggle to capture implicit semantic correlations between image and text features. 
(3) The total local-level objective simply accumulates all localized contrastive losses, implying each local text feature to be treated equally. However, different sentences within medical reports hold varying levels of importance for diagnosis as illustrated in Fig.~\ref{fig:fine_grained}. 

Motivated by the above-mentioned limitations,~\citet{muller2022joint} proposed an improved semantic-driven contrastive method, LoVT, covering both text-to-image local alignment and image-to-text local alignment. It still takes image sub-region features as local image features but divides medical reports into sentences instead of words.
To better capture the implicit semantic features, both text-to-image local features and image-to-text local features are learned by transformer layers instead of weighted summation.
Moreover, the weight for each word's localized contrastive loss is assigned adaptively via transformer's attention map~\citep{dosovitskiy2020image}.
~\cite{cheng2023prior} extend the LoVT by incorporating a conditional reconstruction task for image and text representations. This extension facilitates cross-modality feature interaction and learns more fine-grained scale alignment. Additionally, they propose a prototype memory bank for sentence-level embeddings, expecting to learn high-level text features in the joint image-text space. 
In a parallel study~\citep{zhang2023multi}, a similar methodology is also employed. However, it focuses on reconstructing raw text reports instead of using a prototype memory bank.

Besides this line of research, there are many other studies focusing on multi-scale contrast. 
\citet{liao2021multimodal} optimize the estimated mutual information between local image features and sentence-level text representations to implement local feature alignment. 
\citet{seibold2022breaking} assume that each sentence could convey distinct information for diagnosis, and proposed to perform image-sentence alignment. 
~\citet{palepu2023tier} proposes to penalize the entropy of the text-token to image-patch similarity scores. 

\textcolor{highlight}{
\textbf{Multi-scale contrast in volumetric imaging.}
While the aforementioned studies predominantly focus on chest X-rays, recent research has also investigated volumetric imaging modalities. Modalities such as CT are characterized by their detailed anatomical structures and substantial redundant information. Consequently, recent studies~\citep{ketabi2024tumor,lin2024ct,shui2025largescale,lai2025bridged} tend to utilize semantic-driven multi-scale contrast. This approach is preferred over the methodologies discussed in Sections~\ref{sec: data-efficient} and \ref{sec: knowledge}. Specifically, the fVLM model presented by~\citet{shui2025largescale} performs anatomy-level image-text alignment by identifying distinct anatomical structures using TotalSegmentator~\citep{wasserthal2023totalsegmentator}. This highlights the potential for distilling segmentation or detection capabilities from specialized models.
}

\begin{table*}[!h]\small
    \centering
    \caption{Overview of representative studies focusing on improving CLIP pre-training framework. CLS: classification; ZSC: zero-shot classification; SEG: segmentation; DET: detection; RET: retrieval; VQA: visual question answering; PG: phrase grounding; RG: report generation; ITC: image-text classification; SP: survival prediction.}
    \label{tab:pre-training-paper}
    \resizebox{0.95\textwidth}{!}{
    \begin{tabular}{llp{2cm}p{3cm}p{10cm}} 
    \toprule
      \vcell{Method} & \vcell{Domain} & \vcell{Taxonomy} &\vcell{Evaluation tasks}& \vcell{Highlights} \\[-\rowheight]
    \printcelltop & \printcelltop & \printcelltop & \printcelltop  & \printcelltop \\ 
    \midrule
    \vcell{GLoRIA~\citep{huang2021gloria}} & 
    \vcell{Chest X-ray} 
     & 
    \vcell{Multi-scale} & 
    \vcell{CLS, SEG, \newline ZSC,  RET} &
    \vcell{
    GLoRIA jointly learned global and local image-text representations by contrasting attention-weighted image regions with words in the paired reports.
    }
     \\[-\rowheight]
    \printcelltop & \printcelltop & \printcelltop & \printcelltop & \printcelltop \\ 
    \midrule
    \vcell{LocalMI~\citep{liao2021multimodal}} 
    & 
    \vcell{Chest X-ray} 
    & 
    \vcell{Multi-scale} 
    \vcell{CLS} &
    & 
    \vcell{
    This study proposed to estimate and optimize mutual information between local image representation and sentence-level text representation.
    } \\[-\rowheight]
    \printcelltop & \printcelltop & \printcelltop & \printcelltop & \printcelltop\\
    \midrule
    \vcell{LoVT~\citep{muller2022joint}} & \vcell{Chest X-ray} & 
    \vcell{Multi-scale} & 
    \vcell{DET, SEG} &
    \vcell{
    LoVT leveraged self-attention to align sentence-level text representation and patch-level image representation. 
    It adaptively weighed local representations via transformer's attention map.
    } \\[-\rowheight]
    \printcelltop & \printcelltop & \printcelltop & \printcelltop & \printcelltop\\
    \midrule
    \vcell{PRIOR~\citep{cheng2023prior}} & 
    \vcell{Chest X-ray} & 
    \vcell{Multi-scale} & 
    \vcell{CLS, SEG, DET, \newline ZSC, RET} &
    \vcell{
    To realize more fine-grained alignment, a cross-modality conditional reconstruction module was proposed for masked image modeling and sentence prototype generation.
    } \\[-\rowheight]
    \printcelltop & \printcelltop & \printcelltop & \printcelltop & \printcelltop\\
    \midrule
    \vcell{~\cite{muller2022role}} 
    & 
    \vcell{Chest X-ray} & \vcell{Multi-scale} & 
    \vcell{DET, SEG} &
    \vcell{
    It analyzed from the view of distribution prior and argued that global-level and local-level alignment act complementarily. A local uniformity loss was hence proposed to replace local alignment.
    } \\[-\rowheight]
    \printcelltop & \printcelltop & \printcelltop & \printcelltop & \printcelltop\\
    \midrule
    \vcell{MRM~\citep{zhou2023advancing}} 
    & 
    \vcell{Chest X-ray} & \vcell{Multi-scale} & 
    \vcell{CLS, SEG } &
    \vcell{
    Masked image reconstruction and report completion acted as two complementary objectives during pre-training.
    } \\[-\rowheight]
    \printcelltop & \printcelltop & \printcelltop & \printcelltop &\printcelltop\\
    \midrule
    \vcell{\textcolor{highlight}{fVLM}~\citep{shui2025largescale}} 
    & 
    \vcell{\textcolor{highlight}{Chest CT}} & \vcell{\textcolor{highlight}{Multi-scale}} & 
    \vcell{\textcolor{highlight}{CLS, SEG, RG} } &
    \vcell{
    \textcolor{highlight}{They leveraged TotalSegmentator~\citep{wasserthal2023totalsegmentator} to generate masks for 104 pre-defined anatomical structures within CT scans. Correspondingly, each CT report was decomposed into several anatomy-level descriptions, allowing for anatomy-level image-text alignment.}
    } \\[-\rowheight]
    \printcelltop & \printcelltop & \printcelltop & \printcelltop &\printcelltop\\
    \midrule
    \vcell{\textcolor{highlight}{BrgSA}~\citep{lai2025bridged}} 
    & 
    \vcell{\textcolor{highlight}{Chest CT}} & \vcell{\textcolor{highlight}{Multi-scale\newline Knowledge}} & 
    \vcell{\textcolor{highlight}{ZSC, SEG, RET} } &
    \vcell{
    \textcolor{highlight}{BrgSA adopted a large language model to extract useful information from reports in an offline manner. It then designed a cross-modal knowledge interaction module to facilitate interaction between image and text.}
    } \\[-\rowheight]
    \printcelltop & \printcelltop & \printcelltop & \printcelltop &\printcelltop\\
    \midrule
    \vcell{~\cite{pan2022vision}} 
    & 
    \vcell{Placenta} & \vcell{Multi-scale} & 
    \vcell{CLS} &
    \vcell{
    To address the issue of feature suppression during contrastive alignment, this study proposed a NegLogCosh similarity to replace cosine similarity.
    } \\[-\rowheight]
    \printcelltop & \printcelltop & \printcelltop & \printcelltop &\printcelltop\\
    \midrule
    \vcell{~\cite{pan2023enhancing}} 
    & 
    \vcell{Placenta} & \vcell{Multi-scale} & 
    \vcell{CLS} &
    \vcell{
    A Distributional Feature Recomposition (DFR) module was proposed to estimate the significance of each local text feature in a distribution-aware manner.
    } \\[-\rowheight]
    \printcelltop & \printcelltop & \printcelltop & \printcelltop &\printcelltop\\
    \midrule
    \vcell{TCSA~\citep{lei2023tcsa}} 
    & 
    \vcell{Chest X-ray} & \vcell{Multi-scale\newline Data-efficient} & 
    \vcell{CLS, ZSC, RET} &
    \vcell{
    Besides global-local image-text contrast, TCSA learned view-speciﬁc latent space for multi-view samples and then projected them to a common latent space, which enabled the effective utilization of multi-view information.
    } \\[-\rowheight]
    \printcelltop & \printcelltop & \printcelltop & \printcelltop &\printcelltop\\
    \midrule
    \vcell{MedCLIP~\citep{wang2022medclip}} 
    & 
    \vcell{Chest X-ray} & \vcell{Data-efficient} & 
    \vcell{CLS, ZSC, RET} &
    \vcell{
    It employed inter-report semantical correlation as the soft optimization
target for the alignment between image and text.
    } \\[-\rowheight]
    \printcelltop & \printcelltop & \printcelltop & \printcelltop &\printcelltop\\
    \midrule
    \vcell{SAT~\citep{liu2023improving}} 
    & 
    \vcell{Chest X-ray} & \vcell{Multi-scale\newline Data-efficient} & 
    \vcell{CLS, SEG, \newline ZSC, RET } &
    \vcell{
    SAT categorized image-text pairs into positive/neutral/negative pairs given inter-report similarity and expected to more precisely alleviate false-negative problems. This technique can serve as a plug-and-play module for fine-grained contrast methods.
    } \\[-\rowheight]
    \printcelltop & \printcelltop & \printcelltop & \printcelltop &\printcelltop\\
    \midrule
    \vcell{UMCL~\citep{wang2023umcl}} 
    & 
    \vcell{Chest X-ray} & \vcell{Data-efficient} & 
    \vcell{CLS, ZSC, RET } &
    \vcell{
    It constructed positive/negative pairs through the multi-hot annotation to eliminate false negative pairs.
    } \\[-\rowheight]
    \printcelltop & \printcelltop & \printcelltop & \printcelltop &\printcelltop\\
    \midrule
    \vcell{CXR-CLIP~\citep{you2023cxr}} 
    & 
    \vcell{Chest X-ray} & \vcell{Data-efficient} & 
    \vcell{CLS, ZSC, RET } &
    \vcell{
    CXR-CLIP generated prompts based on ground-truth labels to
provide supplementary information for alignment.
    } \\[-\rowheight]
    \printcelltop & \printcelltop & \printcelltop & \printcelltop &\printcelltop\\
    \midrule
    \vcell{~\cite{jang2022significantly}} 
    & 
    \vcell{Chest X-ray} & \vcell{Data-efficient} & 
    \vcell{ZSC } &
    \vcell{
    This study introduced a sentence-level text augmentation technique and proposed a novel similarity function to compel the model to concentrate on perfectly negative samples instead of false-negative pairs.
    } \\[-\rowheight]
    \printcelltop & \printcelltop & \printcelltop & \printcelltop &\printcelltop\\
    \midrule
    \vcell{MGCA~\citep{wang2022multi}} 
    & 
    \vcell{Chest X-ray} & \vcell{Data-efficient} & 
    \vcell{CLS, DET, SEG} &
    \vcell{
    MGCA sought to leverage disease-level semantic information to cluster samples with high semantic correlation.
    } \\[-\rowheight]
    \printcelltop & \printcelltop & \printcelltop & \printcelltop &\printcelltop\\
    \midrule
    \vcell{BioViL~\citep{boecking2022making}} 
    & 
    \vcell{Chest X-ray} & \vcell{Data-efficient} & 
    \vcell{CLS, SEG,\newline ZSC, PG} &
    \vcell{
    Text data augmentation was adopted to boost training efficiency. 
    Optimized using global alignment alone, BioViL still showed impressive phrase grounding performance, compared with the methods adopting multi-scale contrast.
    } \\[-\rowheight]
    \printcelltop & \printcelltop & \printcelltop & \printcelltop & \printcelltop\\
    \midrule
    \vcell{BioViL-T~\citep{bannur2023learning}} 
    & 
    \vcell{Chest X-ray} & \vcell{Data-efficient} & 
    \vcell{CLS, PG, RG} &
    \vcell{
     It leveraged the temporal connectivity commonly present in diagnostic reports, which were usually discarded in previous studies.
    } \\[-\rowheight]
    \printcelltop & \printcelltop & \printcelltop & \printcelltop & \printcelltop\\
    \bottomrule
    \end{tabular}
    }
\end{table*}

\begin{table*}[!h]\small
    \centering
    \renewcommand{\thetable}{2}
    \caption{(continued)}
    \resizebox{0.95\textwidth}{!}{
    \begin{tabular}{llp{2cm}p{3cm}p{10cm}} 
    \toprule
      \vcell{Method} & \vcell{Domain} & \vcell{Taxonomy} &\vcell{Evaluation tasks}& \vcell{Highlights} \\[-\rowheight]
    \printcelltop & \printcelltop & \printcelltop & \printcelltop  & \printcelltop \\ 
    \midrule
    \vcell{SDA-CLIP~\citep{li2023sda}} 
    & 
    \vcell{\textcolor{highlight}{Surgical Video}} & \vcell{Data-efficient} & 
    \vcell{CLS} &
    \vcell{
    It leveraged the Kullback-Leibler divergence to evaluate the correlation. 
    } \\[-\rowheight]
    \printcelltop & \printcelltop & \printcelltop & \printcelltop & \printcelltop\\
    \midrule
    \vcell{\textcolor{highlight}{SurgVLP}~\citep{yuan2023learning}} 
    & 
    \vcell{\textcolor{highlight}{Surgical Video}} & \vcell{\textcolor{highlight}{Data-efficient}} & 
    \vcell{\textcolor{highlight}{ZSC, RET, RG}} &
    \vcell{
    \textcolor{highlight}{A large-scale video-text dataset was constructed from online lectures via acoustic speech recognition systems.}
    } \\[-\rowheight]
    \printcelltop & \printcelltop & \printcelltop & \printcelltop & \printcelltop\\
    \midrule
    \vcell{\textcolor{highlight}{PeskaVLP}~\citep{yuan2024procedureaware}} 
    & 
    \vcell{\textcolor{highlight}{Surgical Video}} & \vcell{\textcolor{highlight}{Data-efficient \newline Knowledge}} & 
    \vcell{\textcolor{highlight}{CLS, ZSC, RET}} &
    \vcell{
    \textcolor{highlight}{PeskaVLP proposed a novel video-language pre-training strategy to facilitate the understanding of spatial-temporal information at the clip level, phase level, and video level.}
    } \\[-\rowheight]
    \printcelltop & \printcelltop & \printcelltop & \printcelltop & \printcelltop\\
    \midrule
    \vcell{\textcolor{highlight}{EchoPrime}~\citep{vukadinovic2024echoprime}} 
    & 
    \vcell{\textcolor{highlight}{Cardiac Ultrasound}} & \vcell{\textcolor{highlight}{Data-efficient}} & 
    \vcell{\textcolor{highlight}{CLS, RET}} &
    \vcell{
    \textcolor{highlight}{A learnable view classifier was added ahead of the image encoder to explicitly introduce view-specific information during alignment}
    } \\[-\rowheight]
    \printcelltop & \printcelltop & \printcelltop & \printcelltop & \printcelltop\\
     \midrule
    \vcell{\textcolor{highlight}{Merlin}~\citep{blankemeier2024merlin}} 
    & 
    \vcell{\textcolor{highlight}{Chest CT}} & \vcell{\textcolor{highlight}{Data-efficient}} & 
    \vcell{\textcolor{highlight}{CLS, ZSC, SEG, RG, RET}} &
    \vcell{
    \textcolor{highlight}{Besides image-text contrastive learning, diagnosis codes were used as labels for the multi-task training of the vision encoder.}
    } \\[-\rowheight]
    \printcelltop & \printcelltop & \printcelltop & \printcelltop & \printcelltop\\
    \midrule
    \vcell{UniBrain~\citep{lei2023unibrain}} 
    & 
    \vcell{Brain MRI} & \vcell{Knowledge \newline Data-efficient} & 
    \vcell{CLS, ZSC} &
    \vcell{
    $\bullet$ Knowledge encoder: MedKEBERT~\citep{gu2021domain}\newline
    $\bullet$ Multiple pre-defined knowledge prompts served as the query for
the diagnosis of brain disease
    } \\[-\rowheight]
    \printcelltop & \printcelltop & \printcelltop & \printcelltop &\printcelltop\\
    \midrule
    \vcell{FLAIR~\citep{silva2025foundation}} 
    & 
    \vcell{\textcolor{highlight}{Fundus}} & \vcell{Knowledge} & 
    \vcell{CLS, ZSC} &
    \vcell{
    $\bullet$ Knowledge encoder: ClinicalBERT~\citep{alsentzer2019publicly}\newline
    $\bullet$ It incorporated domain knowledge explicitly through descriptive textual prompts during both pre-training and zero-shot inference.
    } \\[-\rowheight]
    \printcelltop & \printcelltop & \printcelltop & \printcelltop &\printcelltop\\
    \midrule
    \vcell{\textcolor{highlight}{KeepFIT}~\citep{wu2024mm}} 
    & 
    \vcell{\textcolor{highlight}{Fundus}} & \vcell{\textcolor{highlight}{Knowledge}} & 
    \vcell{\textcolor{highlight}{CLS, ZSC}} &
    \vcell{
    $\bullet$ \textcolor{highlight}{Knowledge encoder: ClinicalBERT}~\citep{alsentzer2019publicly}\newline
    $\bullet$ \textcolor{highlight}{KeepFIT mixed their high-quality data with other public datasets during pre-training. The low-quality text descriptions from other datasets were refined through visual similarity-based retrieval-augmented generation.}
    } \\[-\rowheight]
    \printcelltop & \printcelltop & \printcelltop & \printcelltop &\printcelltop\\
    \midrule
    \vcell{\textcolor{highlight}{FetalCLIP}~\citep{maani2025fetalclip}} 
    & 
    \vcell{\textcolor{highlight}{Fetal Ultrasound}} & \vcell{\textcolor{highlight}{Knowledge}} & 
    \vcell{\textcolor{highlight}{CLS, ZSC, DET, SEG}} &
    \vcell{
    $\bullet$ \textcolor{highlight}{Knowledge encoder: BERT}~\citep{devlin2019bert}\newline
    $\bullet$ \textcolor{highlight}{By leveraging knowledge within fetal ultrasound textbooks and GPT-4O, the pre-trained FetalCLIP showed promising results on several downstream tasks.}
    } \\[-\rowheight]
    \printcelltop & \printcelltop & \printcelltop & \printcelltop &\printcelltop\\
    \midrule
    \vcell{ARL~\citep{chen2022align}} 
    & 
    \vcell{Chest X-ray} & \vcell{Knowledge} & 
    \vcell{VQA, RET, ITC} &
    \vcell{
    $\bullet$ Knowledge encoder: Graph Attention Network~\citep{velivckovic2018graph}.\newline
    $\bullet$ ARL extracted the entity-relation graph of each medical report as subject-level knowledge. Both image and text embeddings were aligned with knowledge embeddings.
    } \\[-\rowheight]
    \printcelltop & \printcelltop & \printcelltop & \printcelltop &\printcelltop\\
    \midrule
    \vcell{MedKLIP~\citep{wu2023medklip}} 
    & 
    \vcell{Chest X-ray} & \vcell{Knowledge} & 
    \vcell{CLS, SEG,\newline ZSC, RG} &
    \vcell{
    $\bullet$ Knowledge encoder: ClinicalBERT~\citep{alsentzer2019publicly}.\newline
    $\bullet$ MedKLIP pre-processed raw reports and extracted medical-related triplets as the subject-level knowledge. Top-\textit{K} most commonly appearing triplets were leveraged as the domain-level knowledge.
    } \\[-\rowheight]
    \printcelltop & \printcelltop & \printcelltop & \printcelltop &\printcelltop\\
    \midrule
    \vcell{KAD~\citep{zhang2023knowledge}} 
    & 
    \vcell{Chest X-ray} & \vcell{Knowledge} & 
    \vcell{ZSC} &
    \vcell{
    $\bullet$ Knowledge encoder: {Fine-tuned} PubMedBERT~\citep{gu2021domain}.\newline
    $\bullet$ A set of disease descriptions was pre-defined as knowledge queries, allowing for the diagnosis of unseen diseases.
    } \\[-\rowheight]
    \printcelltop & \printcelltop & \printcelltop & \printcelltop &\printcelltop\\
    \midrule
    \vcell{KoBo~\citep{zhang2023knowledge}} 
    & 
    \vcell{Chest X-ray} & \vcell{Knowledge\newline Data-efficient} & 
    \vcell{CLS, ZSC, SEG} &
    \vcell{
    $\bullet$ Knowledge encoder: CompGCN with LTE~\citep{zhang2022rethinking}\newline
    $\bullet$ Medical knowledge was incorporated to distinguish noisy negative samples. 
    } \\[-\rowheight]
    \printcelltop & \printcelltop & \printcelltop & \printcelltop &\printcelltop\\
    \midrule
    \vcell{PathOmics~\citep{ding2023pathology}} 
    & 
    \vcell{\textcolor{highlight}{Histology}} & \vcell{Others} & 
    \vcell{SP} &
    \vcell{
    PathOmics explored genotype-phenotype interactions in complex cancer data. It aligned omics tabular data and image patches in a joint latent space via mean squared error instead of commonly adopted cosine similarity. 
    } \\[-\rowheight]
    \printcelltop & \printcelltop & \printcelltop & \printcelltop &\printcelltop\\
    \midrule
    \vcell{CMTA~\citep{zhou2023cross}} 
    & 
    \vcell{\textcolor{highlight}{Histology}} & \vcell{Others} & 
    \vcell{SP} &
    \vcell{
    CMTA focused on the intrinsic cross-modal correlations between genomic profile and pathology image. 
    } \\[-\rowheight]
    \printcelltop & \printcelltop & \printcelltop & \printcelltop &\printcelltop\\
    \midrule
    \vcell{\textcolor{highlight}{~\citet{turgut2025unlocking}}} 
    & 
    \vcell{\textcolor{highlight}{Cardiac MRI, ECG}} & \vcell{Others} & 
    \vcell{\textcolor{highlight}{CLS, SP}} &
    \vcell{
    \textcolor{highlight}{ECG and Cardiac MRI from the same subject were aligned into a shared latent space using CLIP pre-training.} 
    } \\[-\rowheight]
    \printcelltop & \printcelltop & \printcelltop & \printcelltop &\printcelltop\\
    \midrule
    \vcell{SCA-Net~\citep{chen2023surgical}} 
    & 
    \vcell{\textcolor{highlight}{Surgical Video}} & \vcell{Others} & 
    \vcell{RG} &
    \vcell{
    SCA-Net mutually aligned image and text representations with
prototype representations of the other modality.
    } \\[-\rowheight]
    \printcelltop & \printcelltop & \printcelltop & \printcelltop &\printcelltop\\
    \midrule
    \vcell{M-FLAG~\citep{liu2023m}} 
    & 
    \vcell{Chest X-ray} & \vcell{Others} & 
    \vcell{CLS, DET, SEG} &
    \vcell{
    M-FLAG kept the text encoder fixed during pre-training to alleviate the collapsed solution problem. It explicitly optimized the latent geometry towards orthogonality using the proposed optimization objective.
    } \\[-\rowheight]
    \printcelltop & \printcelltop & \printcelltop & \printcelltop &\printcelltop\\
    \midrule
    \vcell{{~\cite{liu2023utilizing}}} 
    & 
    \vcell{{Chest X-ray}} & \vcell{{Others}} & 
    \vcell{{CLS, DET, SEG}} &
    \vcell{
    {It claimed to be the first to implement CLIP pre-training completely based on real radiology reports and synthetic chest radiographs. The experimental results supported its viability.}
    } \\[-\rowheight]
    \printcelltop & \printcelltop & \printcelltop & \printcelltop &\printcelltop\\
    \bottomrule
    \end{tabular}
    }
\end{table*}

\subsection{Data-efficient contrast}\label{sec: data-efficient}
Large-scale medical imaging datasets with paired reports are hard to obtain due to ethical concerns, which adversely influences the effectiveness of CLIP pre-training due to its data-thirsty nature~\citep{zhai2022scaling,cherti2023reproducible}. To tackle this challenge, various studies have endeavored to implement contrastive image-text pre-training in a more efficient manner, which broadly falls into two categories, i.e., correlation-driven contrast and data mining.

\textbf{Correlation-driven contrast.} Several studies manage to boost the efficiency of contrastive pre-training based on semantic correlation. 
One notable distinction between medical reports and image captions lies in the fact that medical reports are written with a clear diagnostic purpose. Since a small proportion of diseases or findings typically cover most cases~\citep{bustos2020padchest,zhang2023challenges}, the semantic overlap between medical reports can be significant, especially for normal cases as shown in Fig.~\ref{fig:false_negative}. As a result, simply treating unpaired images and reports as negative pairs can lead to issues of false-negative and degrade the efficacy of pre-training. 
Motivated by this observation, ~\cite{wang2022medclip} construct a correlation matrix for all medical reports within the training set following the practice of NegBio~\citep{peng2018negbio}. 
The pre-computed correlation is then employed as a soft optimization target, instead of the original one-hot optimization target, for image-text alignment, enabling the effective utilization of unpaired false-negative reports.
Under conditions where medical reports are not available, \cite{wang2023umcl} construct multi-hot optimization targets for contrastive pre-training by measuring the correlation between ground-truth labels of different samples. 
In a recent study, \cite{liu2023improving} further categorized all image-text pairs into positive, negative, and neutral pairs based on inter-report correlation. This improved categorization of sample pairs allows for a more precise mining of false-negative pairs. 
MGCA~\citep{wang2022multi} focuses on the disease-level inter-sample correlation, a level of abstraction higher than that of image-level semantics. It has designed a novel cross-modal disease-level alignment framework to serve samples with the same diseases.
The semantic correlation can also be extended to the imaging modality level.
For multi-modal brain MRI and corresponding modality-wise reports, UniBrain~\citep{lei2023unibrain} aligns the modality-wise image-text features and then concatenates these features together to realize subject-wise image-text alignment.

\textbf{Data mining.} Simultaneously, many other studies try to boost training efficiency by mining supplementary information.
Within a diagnostic report, the \textit{Findings} section provides a detailed description of clinical observations, while the \textit{Impression} section typically encapsulates these findings and offers an overall assessment~\citep{wallis2011radiology,ganeshan2018structured}.
While previous studies primarily focused on extracting the \textit{Findings} section from raw diagnostic reports, often overlooking the \textit{Impression} section, ~\cite{boecking2022making} incorporate the latter section to enrich the information available for image-text alignment. 
In addition, since the dependency between sentences is weak (see Fig.~\ref{fig:fine_grained}), they also propose to randomly shuffle sentences within each section. CXR-CLIP~\citep{you2023cxr} has explored and utilized uncertainty annotations~\citep{irvin2019chexpert}. It generates prompts based on uncertainty annotations to provide supplementary information for image-text alignment. 

\textcolor{highlight}{
\textbf{Data-efficient contrast in video-based imaging.} In our review, we identified a line of research focusing specifically on video-based imaging modalities such as endoscopy~\citep{li2023sda}, laparoscopy~\citep{yuan2023learning,yuan2024hecvl,yuan2024procedureaware}, and ultrasound~\citep{christensen2024vision,vukadinovic2024echoprime}. Given their shared clinical context, we here classify endoscopy and laparoscopy under the umbrella of \textit{surgical video}. Both surgical video and ultrasound video contain rich spatio-temporal information, distinguishing them from other modalities and motivating a strong emphasis on data-efficient contrast strategies. ~\citet{yuan2023learning} pioneers the development of the first large-scale video-text dataset in this domain by leveraging online lectures and acoustic speech recognition. Their subsequent studies~\citep{yuan2024hecvl,yuan2024procedureaware} refine image-text alignment methods by leveraging the sequential nature of surgical video. Notably, PeskaVLP~\citep{yuan2024procedureaware} achieves significant progress by integrating large language models for enhanced text supervision and also hierarchically aligning text with video clips, procedural phases, and entire videos.
}

\subsection{Explicit knowledge enhancement}\label{sec: knowledge}

While studies in Section~\ref{sec: global-local} and Section~\ref{sec: data-efficient} in nature still focus on {internal} information of the dataset, some researchers have investigated the integration of {external} medical knowledge to enhance the pre-training process.

Employing the unified medical language system (UMLS) as the external knowledge base~\citep{bodenreider2004unified} for medical concepts, existing studies typically incorporate knowledge enhancement at both the {subject level} and the {domain level} as shown in Fig.~\ref{fig:knowledge-enhanced}.
At the \textbf{subject level}, ScispaCy~\citep{neumann2019scispacy}, a named entity recognition tool, is commonly adopted to extract medical entities from each report and link them to the corresponding medical concepts in the UMLS for entity disambiguation. Then, an entity-relation graph is built based on relations defined in UMLS or RadGraph~\citep{jain2021radgraph}, where the former establishes relations for general medical concepts and the latter is tailored to the practice of chest X-ray. These relations, including causal, positional, and modifying relationships, can provide an illustration of the image's visual structures and the process of human reasoning, enhancing the alignment within each image-text pair.
For \textbf{domain-level} enhancement, knowledge is typically represented as a domain-specific knowledge graph or a descriptive knowledge prompt for the targeted medical imaging domain (e.g., chest X-ray, and brain MRI), covering related organs, tissues, or clinical findings. The domain-specific knowledge graph can either be defined as a trainable symbolic graph, or a set of top-\textit{K} most common entity triplets~\citep{wu2023medklip}. The descriptive knowledge prompt usually provides detailed observations or explanations for encompassed entities.

To incorporate external knowledge into the pre-training phase, an auxiliary \textbf{knowledge encoder} is usually involved. It serves to convert knowledge information into knowledge embeddings that can be interpreted by neural networks. 
The choice of the knowledge encoder may involve the selection of a graph neural network~\citep{chen2022align,zhang2023knowledge} or a pre-trained BERT model~\citep{lin2023towards}.

ARL~\citep{chen2022align} is the representative study of subject-level knowledge enhancement. Specifically, it adopts TransE~\citep{bordes2013translating} algorithm to train a graph attention network~\citep{velivckovic2018graph} as the knowledge encoder. All medical reports in the training set are pre-processed to construct subject-specific entity-relation graphs according to relations defined in the UMLS~\citep{bodenreider2004unified}. 
In contrast, KoBo~\citep{zhang2023knowledge} and FLAIR~\citep{wang2023foundation} prioritize domain-level enhancement. KoBo extracts a knowledge graph containing radiological medical concepts from UMLS and utilizes CompGCN~\citep{zhang2022rethinking} as the knowledge encoder. It has proposed a knowledge semantic enhancement module and a knowledge semantic guidance module to mitigate negative sample noise and adjust semantic shifting, respectively.  Similarly, FLAIR leverages ClinicalBERT~\citep{alsentzer2019publicly} to interpret human-refined knowledge descriptions for the domain of fundus imaging, including detailed descriptions of visual features and inter-concept relationships. 

In addition to the studies mentioned above, there are also studies adopting both subject-level and domain-level knowledge enhancements. 
MedKLIP~\citep{wu2023medklip} has pre-processed and extracted entity triplets from each raw text report, constituting subject-level knowledge. Then, the top-\textit{K} most frequently occurring entities, i.e.,~\textit{K}=75 in practical implementation, are identified to form an entity query set, functioning as the domain-level knowledge graph. The ClinicalBERT~\citep{alsentzer2019publicly} is adopted for subject-level and domain-level knowledge enhancements, simultaneously.
This paradigm is also adopted in KAD~\citep{zhang2023knowledge}, revealing its robustness. MOTOR~\citep{lin2023towards} adopts a symbolic graph regarding key clinical findings as the domain-level knowledge and also extracts entity-relation graphs for subject-level enhancement. The pre-trained SciBERT~\citep{beltagy-etal-2019-scibert} is adopted as the knowledge encoder of MOTOR. 
The improved performance indicates the rationality of explicit knowledge enhancement in CLIP-style pre-training.

\begin{table*}[tbp]
\footnotesize
\centering
\renewcommand{\thetable}{3}
\caption{Overview of representative classification applications.}
\label{tab:classification}
\resizebox{1.0\textwidth}{!}{
\begin{tabular}{lllp{5.0cm}l} 
\toprule
\vcell{\textbf{Method}} & \vcell{\textbf{Organ}} &\vcell{\textbf{Modality}} &\vcell{\textbf{Test dataset}} &\vcell{\textbf{Pre-trained CLIP}} \\[-\rowheight]
\printcelltop & \printcelltop & \printcelltop  & \printcelltop  & \printcelltop \\ 
\midrule
\vcell{CheXzero~\citep{tiu2022expert}} & 
\vcell{Chest}& 
\vcell{X-ray}&
\vcell{CheXpert~\citep{irvin2019chexpert}\newline PadChest~\citep{bustos2020padchest}} & 
\vcell{Fine-tuned CLIP~\citep{radford2021learning}} 
\\[-\rowheight]
\printcelltop & \printcelltop& \printcelltop& \printcelltop   & \printcelltop \\ 
\midrule
\vcell{Xplainer~\citep{pellegrini2023xplainer}} & 
\vcell{Chest} & 
\vcell{X-ray}&
\vcell{CheXpert~\citep{irvin2019chexpert}\newline ChestX-ray14~\cite{wang2017chestx}} & 
\vcell{BioViL~\citep{boecking2022making}} 
\\[-\rowheight]
\printcelltop  & \printcelltop& \printcelltop& \printcelltop   & \printcelltop \\ 
\midrule
\vcell{~\cite{kumar2022towards}} & 
\vcell{Chest} & 
\vcell{X-ray}&
\vcell{MIMIC-CXR~\citep{johnson2019mimic}
} & 
\vcell{Fine-tuned BioViL~\citep{boecking2022making}} 
\\[-\rowheight]
\printcelltop  & \printcelltop& \printcelltop& \printcelltop   & \printcelltop \\ 
\midrule
\vcell{~\cite{pham2024ai}} & 
\vcell{Chest} & 
\vcell{X-ray}&
\vcell{EGD-CXR~\citep{karargyris2021creation}
} & 
\vcell{BiomedCLIP~\citep{zhang2023large}} 
\\[-\rowheight]
\printcelltop  & \printcelltop& \printcelltop& \printcelltop   & \printcelltop \\ 
\midrule
\vcell{~\cite{mishra2023improving}} & 
\vcell{Chest} & 
\vcell{X-ray}&
\vcell{VinDr-CXR~\citep{nguyen2022vindr}} & 
\vcell{Fine-tuned CLIP~\citep{radford2021learning}} 
\\[-\rowheight]
\printcelltop  & \printcelltop& \printcelltop& \printcelltop   & \printcelltop \\
\midrule
\vcell{ETP~\citep{liu2024etp}} & 
\vcell{Heart} & 
\vcell{ECG}&
\vcell{PTB-XL~\citep{wagner2020ptb}\newline CPSC2018~\citep{liu2018open}} & 
\vcell{An ECG CLIP trained from scratch} 
\\[-\rowheight]
\printcelltop  & \printcelltop& \printcelltop& \printcelltop   & \printcelltop \\
\midrule
\vcell{CITE~\citep{zhang2023text}} & 
\vcell{Stomach} & 
\vcell{Histology}&
\vcell{PatchGastric~\citep{tsuneki2022inference}} & 
\vcell{CLIP~\citep{radford2021learning}} 
\\[-\rowheight]
\printcelltop  & \printcelltop& \printcelltop& \printcelltop   & \printcelltop \\
\midrule
\vcell{CLIP-Lung~\citep{CLIP-lung_coop}} & 
\vcell{Lung} & 
\vcell{CT}&
\vcell{LIDC-IDRI~\citep{armato2011lung}} & 
\vcell{CLIP~\citep{radford2021learning}} 
\\[-\rowheight]
\printcelltop  & \printcelltop& \printcelltop& \printcelltop   & \printcelltop \\
\midrule
\vcell{~\cite{liu2024progressive}} & 
\vcell{Brain} & 
\vcell{MRI}&
\vcell{ADNI~\citep{jack2015magnetic}\newline OASIS~\citep{lamontagne2019oasis}\newline NACC~\citep{beekly2007national}} & 
\vcell{BiomedCLIP~\citep{zhang2023large}} 
\\[-\rowheight]
\printcelltop  & \printcelltop& \printcelltop& \printcelltop   & \printcelltop \\ 
\midrule
\vcell{DCPL~\citep{cao2024domain}} & 
\vcell{Brain, Colorectal} & 
\vcell{MRI, Histology}&
\vcell{BTMRI~\citep{nickparvar_2021}\newline CHMNIST~\citep{kather2016multi}} & 
\vcell{CLIP~\citep{radford2021learning}} 
\\[-\rowheight]
\printcelltop  & \printcelltop& \printcelltop& \printcelltop   & \printcelltop \\ 
\midrule
\vcell{CoOPLVT~\citep{baliah2023exploring_coop}} & 
\vcell{Eye} & 
\vcell{Fundus}&
\vcell{EyePACS~\citep{diabetic-retinopathy-detection}\newline APTOS~\citep{aptos2019-blindness-detection}\newline 
 Messidor~\citep{decenciere2014feedback}
} & 
\vcell{CLIP~\citep{radford2021learning}} 
\\[-\rowheight]
\printcelltop  & \printcelltop& \printcelltop& \printcelltop   & \printcelltop \\ 
\midrule
\vcell{~\cite{byra2023few_llm_gen_pro}} & 
\vcell{Chest, Breast} & 
\vcell{X-ray, Ultrasound}&
\vcell{Pneumonia~\citep{kermany2018identifying}\newline UDIAT~\citep{yap2017automated}
} & 
\vcell{CLIP~\citep{radford2021learning}} 
\\[-\rowheight]
\printcelltop  & \printcelltop& \printcelltop& \printcelltop   & \printcelltop \\ 
\bottomrule 
\end{tabular}}

\end{table*}

\subsection{Others}\label{sec: pre-train-summary}

While existing methods can generally be categorized according to the taxonomy previously outlined, exceptions still exist. 
These exceptions may provide insight for potential research.
M-FLAG~\citep{liu2023m} focuses on the issue of collapse solution~\citep{grill2020bootstrap,chen2021exploring},
{which means that image and text information are falsely encoded into constant feature embeddings to ensure a near-zero distance in the latent space.}
To address this issue, they propose to keep the text encoder frozen during pre-training, and adopt an orthogonality loss to encourage the orthogonality of visual representations.
{~\cite{liu2023utilizing} report to be the first attempt to implement CLIP pre-training based on synthetic chest x-rays and paired real radiology reports. Their experimental results revealed the potential value of domain-specific generative
models~\citep{chambon2022roentgen} in the medical imaging community.}
CMTA~\citep{zhou2023cross} and PathOmics~\citep{ding2023pathology} investigate the alignment between omics tabular data and pathology images, which could potentially inspire the exploration of aligning other forms of data with images beyond diagnostic reports. 
\textcolor{highlight}{It is worth noting that~\cite{turgut2025unlocking} introduce an innovative approach by applying CLIP-style pre-training to align two distinct imaging modalities: electrocardiograms and cardiac MRI. While electrocardiograms are low-price, they have limited diagnostic capabilities. While cardiac MRIs can provide more definitive diagnostic information, they are much more expensive. This method successfully unlocks the potential of electrocardiograms in diagnosing cardiovascular diseases, offering a low-cost and effective choice for patients. We believe this recent work can offer valuable insights and benefit the broader research community.}

\subsection{Summary}\label{sec: pre-train-gt-summary}
In this section, we provide an overview of adapting CLIP pre-training for medical imaging, which is taxonomized into three categories.
Despite their categorization, these methods are inherently the same -- they all attempt to explore and utilize the inherent consistency relationships within paired images and text.
The multi-scale contrast approach focuses on consistency between local image features and text features, enabling more detailed interpretation. Simultaneously, the data-efficient contrast approach emphasizes the efficient use of data, maintaining consistency across different samples by leveraging inter-sample correlations. Lastly, the knowledge-enhanced methods {utilize expert-level knowledge to explicitly interpret the inherent relationships between medical entities, going beyond basic image-text matching to ensure that the learned image and text representations are in line with the nuanced and complex knowledge of medical experts.} Each of these methods contributes uniquely to the medical imaging domain, showcasing the adaptability and potential of CLIP pre-training to \textit{not only} enhance traditional image-text relationships \textit{but also} introduce a new depth and precision in medical image analysis.

\section{CLIP-driven applications}\label{sec: downstream}
Leveraging large-scale text supervision, the pre-trained CLIP model effectively aligns visual features with human language, a capability that extends to medical images (referring to Fig.~\ref{fig:clip-explainable}). This capability is particularly significant in clinical settings where interpretability is of importance~{\citep{lauritsen2020explainable,tjoa2020survey}}. Meanwhile, the rich human knowledge embedded in CLIP can also act as an external supervision for annotation-demanding tasks, such as tumor segmentation~\citep{liu2023clip}. These strengths of CLIP explain the growing adoption of CLIP in various clinical applications.
\begin{figure}[tbp]
    \centering
    \includegraphics[width=0.45\textwidth]{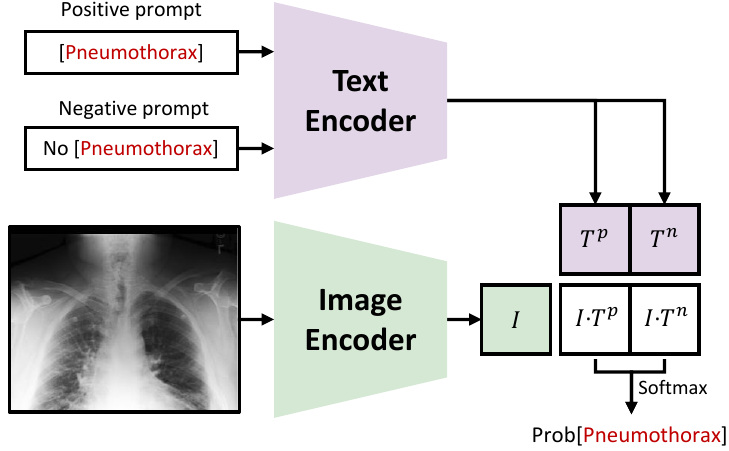}
    \caption{Illustration of the positive/negative prompt engineering for zero-shot disease diagnosis. The diagnosis of Pneumothorax is demonstrated here, while other potential diseases can also be diagnosed in this way.}
    \label{fig:clip-zero-shot}
\end{figure}
\begin{figure*}[t]
    \centering
    \includegraphics[width=1.0\textwidth]{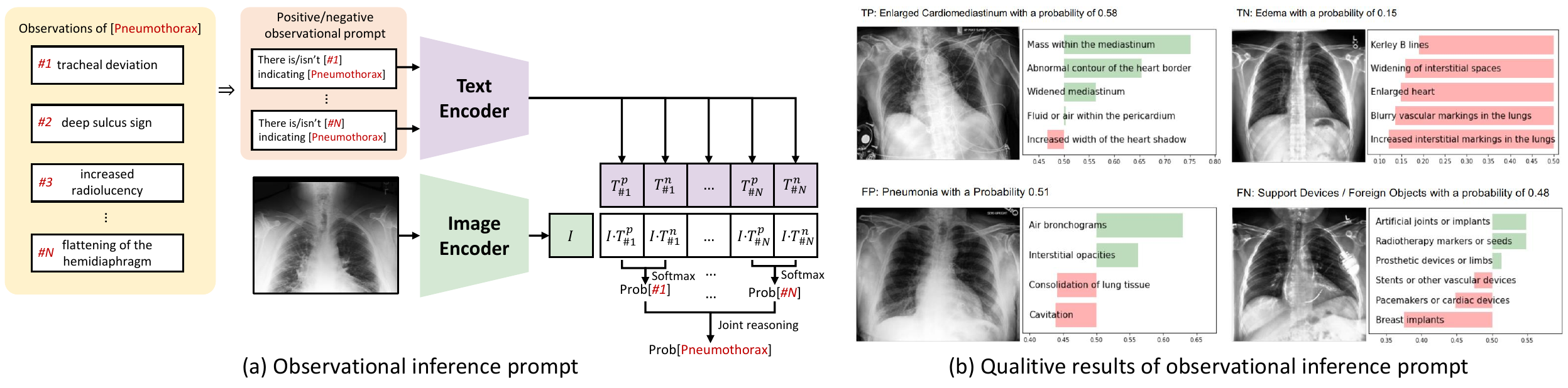}
    \caption{Overview of Xplainer. (a) Illustration of the observational inference prompt proposed by Xplainer. The diagnosis of Pneumothorax is demonstrated here, while other potential diseases can also be implemented in this way. (b) The qualitative results of Xplainer.}
    \label{fig:xplainer_results}
\end{figure*}
\subsection{Classification}
The pre-training of CLIP involves image-text alignment, making it a natural fit for medical image classification~\citep{xian2016latent,radford2021learning}. This task typically requires a global assessment of the image (e.g., determining whether it is benign or malignant or identifying specific diseases). In Table~\ref{tab:classification}, we present existing studies that employ CLIP for image classification, which can generally be categorized into two approaches: zero-shot classification and context optimization.
\textbf{Zero-shot classification} focuses on leveraging the diagnostic potential of a domain-specific CLIP model through prompt engineering.
\textbf{Context optimization}, on the other hand, aims to fine-tune a non-domain-specific CLIP model to the medical domain in a manner that is both parameter- and data-efficient.

\subsubsection{Zero-shot classification}
This research avenue attempts to fully leverage the potential of pre-trained CLIP models without fine-tuning on downstream datasets. Hence, the diagnostic performance would largely depend on the built-in knowledge, which influences the choice of domain-specific CLIP.
In existing studies, a domain-specific CLIP is typically obtained by either fine-tuning the original CLIP on the target medical imaging domain~\citep{tiu2022expert,seibold2022breaking,mishra2023improving,kumar2022towards} or by adopting an open-source specialized CLIP model~\citep{boecking2022making,pham2024ai}. For unique cases like electrocardiogram (ECG)~\citep{liu2024etp}, where the data exists in the form of one-dimensional multi-channel signals, researchers often train ECG-specific CLIP models from scratch.

Besides the choice of pre-trained CLIP, the other key point of zero-shot classification lies in prompt engineering~{\citep{wei2022chain,wang2023selfconsistency,gu2023systematic}}. 
While the standard zero-shot prompt described in Section~\ref{clip_theory} is effective for tasks like {BI-RADS grading~\citep{liberman2002breast} or OA assessment~\citep{kellgren1957radiological}}, it falls short in disease diagnosis~\citep{tiu2022expert}. The shortfall is attributed to the softmax operations in the probability calculations (see Eq.~\ref{eq:prob}), which treat each class as mutually exclusive. This does not align with the reality of disease diagnosis, as patients may simultaneously suffer from multiple diseases.
To address this issue, CheXzero~\citep{tiu2022expert} defines the positive and negative prompts (such as ‘Pneumothorax’ versus ‘no Pneumothorax’) to implement zero-shot disease diagnosis for different potential diseases in a compatible manner (see Fig.~\ref{fig:clip-zero-shot}). 
\begin{figure}[tbp]
    \centering
    \includegraphics[width=0.5\textwidth]{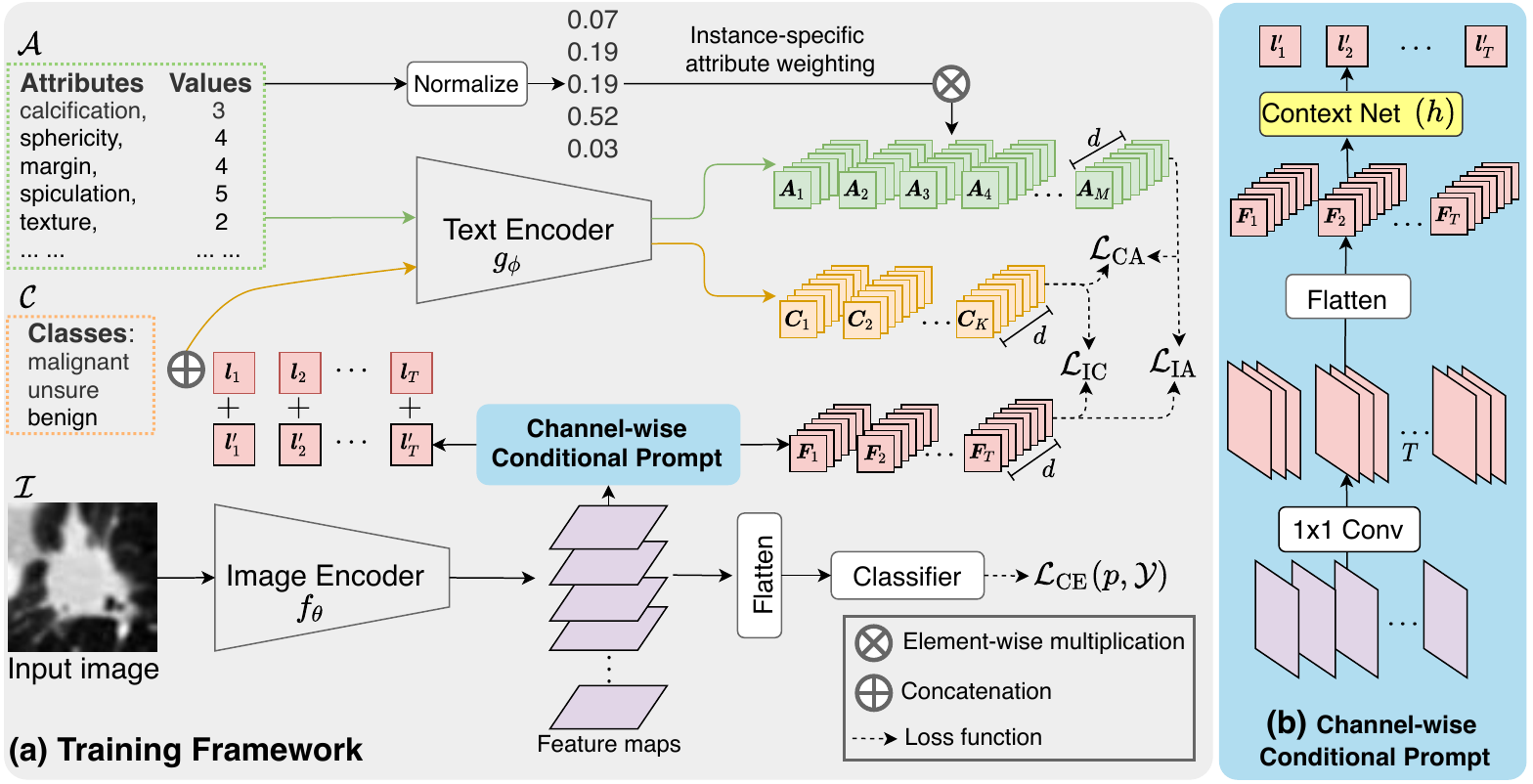}
    \caption{Context optimization for lung nodule classification (from ~\cite{CLIP-lung_coop}).}
    \label{fig:clip-CoOp}
\end{figure}

\begin{figure*}[tbp]
    \centering.
\includegraphics[width=0.95\textwidth]{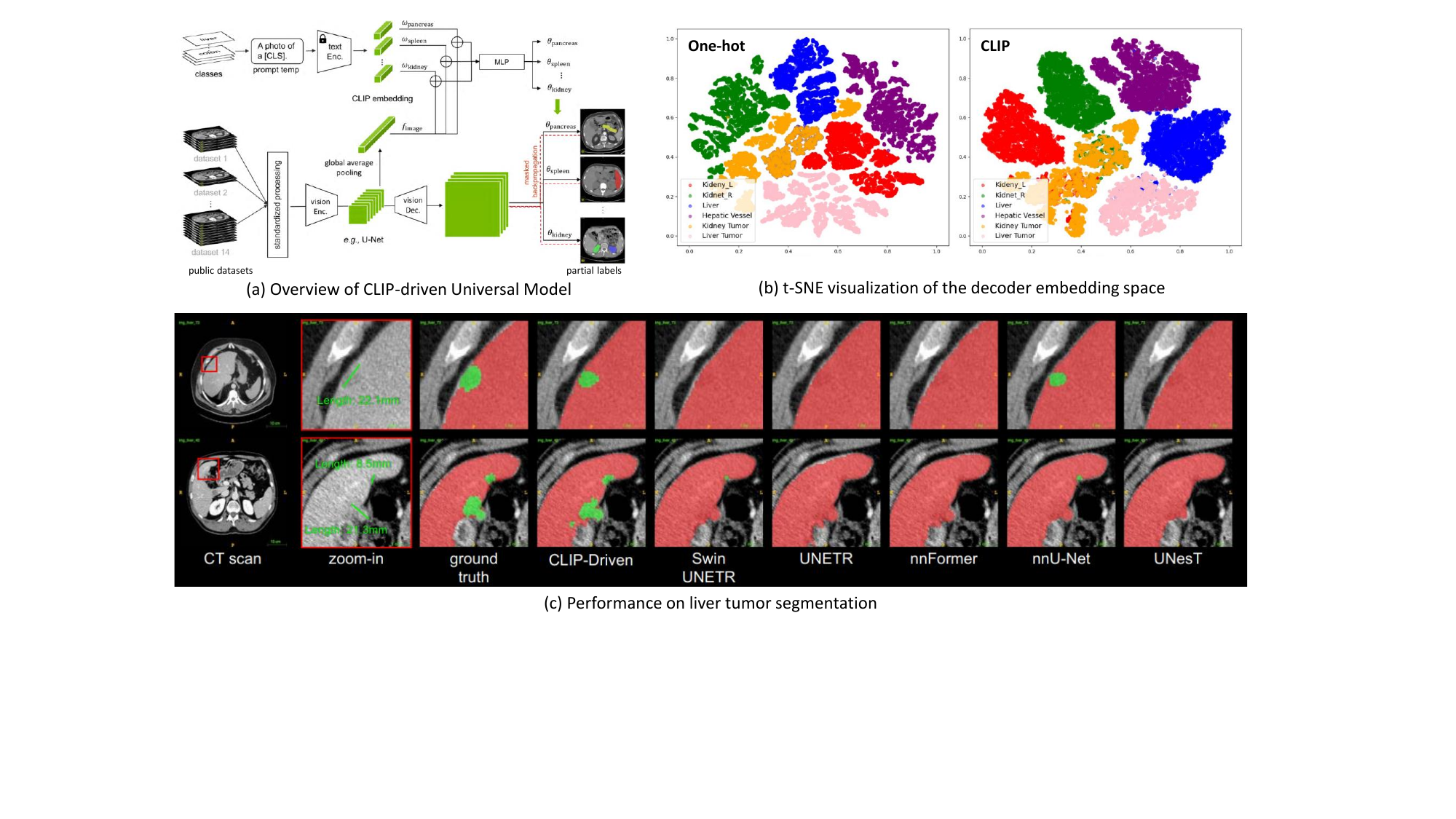}
    \caption{(a) Overview of CLIP-driven segmentation model for universal segmentation\citep{liu2023clip}. (b) t-SNE visualization of the decoder embedding space between one-shot task encoding and CLIP label encoding. (c) Performance on liver tumor segmentation (green for tumor and red for organ).}
    \label{fig:clipuniseg}
\end{figure*}

However, the diagnosis provided by CheXzero lacks explainability. To alleviate this issue, several methods have tried to incorporate detailed descriptions of the image in the prompt design, including texture, shape, and location.~\citep{byra2023few_llm_gen_pro, kim2023concept_llm_gen_pro, yan2023robust_llm_gen_pro, liu2023chatgpt_llm_gen_pro, pellegrini2023xplainer}.
Among them, \cite{pellegrini2023xplainer} introduces Xplainer, a representative method for explainable zero-shot diagnosis. Specifically, instead of directly predicting a diagnosis, they prompt the model to classify the existence of descriptive observations, for which a radiologist would look for on an X-ray scan, and use the joint probabilities of all observations to estimate the overall probability. 
These descriptive prompts are originally generated by ChatGPT by querying observations that may indicate specific diseases on a Chest X-ray. To improve the reliability of these prompts, radiologists are asked to further refine them based on their experience.
During the diagnosis of a specific disease (e.g., Pneumothorax), 
all related observations would be fed into the CLIP text encoder in the form of a positive/negative prompt, as shown in Fig. \ref{fig:xplainer_results}(a).
According to the probability of each observation (from Prob[\#1] to Prob[\#\textit{N}]), a joint probability can be estimated as the final result. 
Fig.~\ref{fig:xplainer_results}(b) gives a qualitative demonstration of Xplainer's explainable diagnosis.
It can be observed that Xplainer can correctly detect true positive and true negative cases.
Although it fails to always make correct decisions in false positive and false negative cases, both of them show contradictory findings (e.g., bronchogram tends to co-occur with consolidation), which means they can be easily detected and corrected by radiologists. The illustration of underlying reasons in Xplainer would no doubt improve the explainability of zero-shot diagnosis. 

\begin{table*}[ht!]
\footnotesize
\centering
\renewcommand{\thetable}{4}
\caption{Overview of representative dense prediction applications.}
\label{tab:dense_predictions}
\begin{tabular}{llp{2.5cm}p{3cm}p{4.1cm}}
\toprule 
    \vcell{\textbf{Method}} &\vcell{\textbf{Task}}& \vcell{\textbf{Organ}} &\vcell{\textbf{Modality}} &\vcell{\textbf{Pre-trained CLIP}} \\[-\rowheight]
\printcelltop & \printcelltop & \printcelltop  & \printcelltop  & \printcelltop \\ 
\midrule
\vcell{\cite{guo2023multiple}} & 
\vcell{Detection}& 
\vcell{\textcolor{highlight}{General}}&
\vcell{\textcolor{highlight}{General}} & 
\vcell{GLIP~\citep{li2022grounded}} 
\\[-\rowheight]
\printcelltop & \printcelltop& \printcelltop& \printcelltop   & \printcelltop \\
\midrule
\vcell{VLPMNuD~\citep{wu2023zero}} & 
\vcell{Detection}& 
\vcell{Nuclei}&
\vcell{Cytology} & 
\vcell{GLIP~\citep{li2022grounded}} 
\\[-\rowheight]
\printcelltop & \printcelltop& \printcelltop& \printcelltop   & \printcelltop \\
\midrule
\vcell{AnomalyCLIP~\citep{zhou2024anomalyclip}} & 
\vcell{Anomaly Detection}& 
\vcell{\textcolor{highlight}{General}}&
\vcell{\textcolor{highlight}{General}} & 
\vcell{CLIP~\citep{radford2021learning}} 
\\[-\rowheight]
\printcelltop & \printcelltop& \printcelltop& \printcelltop   & \printcelltop \\
\midrule
\vcell{\textcolor{highlight}{MediCLIP}~\citep{zhou2024anomalyclip}} & 
\vcell{\textcolor{highlight}{Anomaly Detection}}& 
\vcell{\textcolor{highlight}{Brain, Breast, Chest}}&
\vcell{\textcolor{highlight}{MRI, Ultrasound, X-ray}} & 
\vcell{\textcolor{highlight}{CLIP}~\citep{radford2021learning}} 
\\[-\rowheight]
\printcelltop & \printcelltop& \printcelltop& \printcelltop   & \printcelltop \\
\midrule
\vcell{SyntheticBoost~\citep{adhikari2023synthetic}} & 
\vcell{Segmentation}& 
\vcell{Heart}&
\vcell{Ultrasound} & 
\vcell{Fine-tuned CLIPSeg\newline~\citep{luddecke2022image}\newline Fine-tuned CRIS\newline~\citep{wang2022cris}}
\\[-\rowheight]
\printcelltop & \printcelltop& \printcelltop& \printcelltop   & \printcelltop \\
\midrule
\vcell{\textcolor{highlight}{KGPL}~\citep{teng2024knowledge}} & 
\vcell{\textcolor{highlight}{Segmentation}}& 
\vcell{\textcolor{highlight}{Brain}}&
\vcell{\textcolor{highlight}{MRI}} & 
\vcell{\textcolor{highlight}{BiomedCLIP}~\citep{zhang2023large}} 
\\[-\rowheight]
\printcelltop & \printcelltop& \printcelltop& \printcelltop   & \printcelltop \\
\midrule
\vcell{\cite{liu2023clip}} & 
\vcell{Segmentation}& 
\vcell{Abdominal organ}&
\vcell{CT, MRI} & 
\vcell{CLIP~\citep{radford2021learning}} 
\\[-\rowheight]
\printcelltop & \printcelltop& \printcelltop& \printcelltop   & \printcelltop \\
\midrule
\vcell{\cite{zhang2023continual}} & 
\vcell{Segmentation}& 
\vcell{Abdominal organ}&
\vcell{CT} & 
\vcell{CLIP~\citep{radford2021learning}} 
\\[-\rowheight]
\printcelltop & \printcelltop& \printcelltop& \printcelltop   & \printcelltop \\
\midrule
\vcell{TPRO~\citep{zhang2023tpro}} & 
\vcell{Segmentation}& 
\vcell{Breast, Lung}&
\vcell{Histology} & 
\vcell{MedCLIP~\citep{medpixclip}} 
\\[-\rowheight]
\printcelltop & \printcelltop& \printcelltop& \printcelltop   & \printcelltop \\
\midrule
\vcell{TCEIP~\citep{yang2023tceip}} & 
\vcell{Keypoints Localization}& 
\vcell{Tooth}&
\vcell{CBCT slices} & 
\vcell{CLIP~\citep{radford2021learning}} 
\\[-\rowheight]
\printcelltop & \printcelltop& \printcelltop& \printcelltop   & \printcelltop \\
    \bottomrule 
\end{tabular}
\end{table*}
\subsubsection{Context optimization}\label{sec: context_opt}
While the concept of zero-shot disease diagnosis appears impressive and promising, its broader application in the medical imaging community is unfortunately constrained by the limited availability of domain-specific CLIP models{~\citep{zhang2023large}}. 
Consequently, this line of studies has turned to non-domain-specific CLIP models, aiming to efficiently adapt these models to the context of medical imaging domains with optimal use of trainable parameters.

Although parameter-efficient tuning studies~\citep{zhou2022learning,zhou2022conditional,khattak2023maple} have been proposed to adapt CLIP to out-of-distribution natural image datasets, none of them considers the medical imaging domain.~{The lack of consideration of domain characteristics may lead to suboptimal performance.} 
To tackle this issue, several studies focusing on context optimization~\citep{cao2024domain, zheng2024exploring, baliah2023exploring_coop,CLIP-lung_coop} have been proposed.
CLIP-Lung~\citep{CLIP-lung_coop} has proposed the channel-wise conditional prompt (CCP) for lung nodule malignancy prediction as shown in Fig.~\ref{fig:clip-CoOp} shows. 
Different from CoCoOp ~\citep{zhou2022conditional}, it constructs learnable prompts based on channel-level information of feature maps. This adaptively learnable prompt successfully leads to improved results.

\subsection{Dense prediction}
{
Dense prediction~\citep{ViTdense,rao2022denseclip,wang2020DenseCL} involves generating outputs 
for every pixel or subset of pixels in an image (see in Table~\ref{tab:dense_predictions}). This task contrasts with classification, where a single label is assigned to the entire image.}
~{Owing to its capability to align with human cognition, CLIP as well as its variants have been applied to a wide variety of dense prediction tasks in medical imaging.}
Methods involved in this line of study typically function as an auxiliary tool, providing clinicians with valuable information (e.g., regions of target organs, potential lesion regions) to support their decision-making.

\subsubsection{Detection}
{
Detection is a crucial prerequisite in many clinical scenarios, such as surgical planning~\citep{bouget2017vision}, pathological diagnosis~\citep{wang2020focalmix,ribli2018detecting}, and post-operative assessment~\citep{lopez2018fully}.}
Previous 
approaches~\citep{baumgartner2021nndetection,ickler2023taming,wittmann2022focused,yuksel2021dental} in medical image detection typically focus on leveraging image-based features extracted through various convolutional neural networks or with transformer-based architectures. These methods, while effective to a certain extent, often struggle with the nuanced and complex nature of medical images, ~{especially in pathology cases where visual cues are subtle or absent~\citep{wittmann2022focused}}. The pipeline for detection tasks in medical images has been significantly influenced by the advancements and integration of visual-language models, such as directly using CLIP~\citep{muller2022radiological} or its extension GLIP~\citep{li2022grounded}. \cite{guo2023multiple} propose a prompt-ensemble technique for the amalgamation of diverse textual descriptions, which fully leverages GLIP's proficiency in interpreting complex medical scenarios. Moreover, VLPMNuD~\citep{wu2023zero} introduces GLIP for zero-shot nuclei detection in H\&E stained images. It has proposed an automated prompt design method and adopted a self-training framework to polish the predicted boxes iteratively.

While object detection focuses on identifying and localizing specific and pre-defined items, such as tumors or fractures, anomaly detection is also a crucial application~{~\citep{fernando2021deep, tschuchnig2022anomaly}} in medical imaging, which aims to identify deviations from the norm. AnomalyCLIP~\citep{zhou2024anomalyclip} showcases CLIP's capabilities in zero-shot anomaly detection across medical domains. AnomalyCLIP employs object-agnostic text prompts to capture the essence of normality and abnormality across various images. This has forced CLIP to pay more attention to abnormal regions rather than to the main objects shown in the image, thereby facilitating more generalized recognition of anomalies compared with previous methods~\citep{zhou2022conditional, sun2022dualcoop,chen2023zero}.

\subsubsection{2D medical image segmentation}
CLIP is originally pre-trained on the 2D image domain via text supervision. Therefore, it can be easily integrated into 2D medical image segmentation with fine-tuning. Following this idea, \citet{muller2022radiological, anand2023oneshot} apply the CLIP pre-trained image encoder across various medical imaging modalities, including X-rays, ultrasound, and CT/MR (by taking 3D data as 2D slices). Their works demonstrate that CLIP's image encoder, originally trained on natural images, can also deliver impressive performance in medical image segmentation tasks. Furthermore, \cite{poudel2024exploring} and ~\cite{adhikari2023synthetic} employ both pre-trained CLIP image and text encoders to construct a vision-language segmentation model, and finetune it to serve 2D medical image segmentation tasks.

\subsubsection{3D medical image segmentation}
A growing number of publicly available datasets~\citep{simpson2019large,heller2019kits19,liu2020ms,bilic2023liver} have allowed researchers to train 3D segmentation models for segmenting anatomical structures and lesions from volumetric imaging data. However, most of these datasets typically only focus on certain organs or anatomical structures while all task-irrelevant organs and tumors are treated as the background, leading to the issue of partial label~\citep{yan2020learning,lyu2021segmentation}. Consequently, it remains a constraint on how to break the barrier of individual datasets and fully leverage existing data cohorts to expand capabilities of segmentation models.

DoDNet~\citep{zhang2021dodnet} is the first segmentation model~{to tackle such partially-labeled problems in medical images, which provides \textit{not only} superior performance against previous multi-head networks~\citep{chen2019med3d,fang2020multi,shi2021marginal}, \textit{but also} a flexible solution to extend to a newly labeled task. DoDNet introduces} a dynamic segmentation head~\citep{tian2020conditional} tailored to specific tasks, with the task represented as a one-hot embedding. However, such label orthogonality encoding ignores the natural semantic relationship between organs and tumors.~{In DoDNet, tasks like liver segmentation and liver tumor segmentation are treated as orthogonal tasks, even though they clearly exhibit a strong correlation.} This limitation is exacerbated as the number of distinct segmentation tasks increases. 
Hence, label orthogonality encoding fails to generalize effectively when the diversity of tasks grows more complex. 
To tackle the aforementioned challenges and limitations, \cite{liu2023clip} propose a CLIP-Driven Universal Segmentation Model (see Fig.~\ref{fig:clipuniseg}(a)) by introducing the text embedding learned from CLIP to replace the one-hot encoding used in DoDNet~\citep{zhang2021dodnet}. Specifically, they have utilized the pre-trained CLIP text encoder to encode task prompts like 'liver', 'liver tumor', 'left kidney',' right kidney', 'hepatic vessel', 'kidney tumor' etc. Such embeddings are then concatenated with the pooled image features to generate the dynamic segmentation head, which is utilized to refine segmentation results after the vision decoder. 
The advantage of CLIP text encoder over DoDNet's one-hot label encoding is shown in Fig.~\ref{fig:clipuniseg}(b). Correlations between organs and tumors are better established with the fixed CLIP label embedding, i.e., the relationship between liver and liver tumor, and the relationship between kidney and kidney tumor. 
This method provides superior performance \textit{not only} in organ segmentation \textit{but also} in more challenging tumor segmentation tasks, surpassing other image-only segmentation models as shown in Fig.~\ref{fig:clipuniseg}(c). The CLIP-Driven Uniserval Model is compared with five vision-only SOTA segmentation methods. By reviewing the segmentation of tumors, the CLIP-Driven segmentation model succeeds in detecting small tumors, even in cases with multiple tiny tumors, which are ignored by most image-only methods. 
\begin{table*}[tbp]
\small
\centering
\renewcommand{\thetable}{5}
\caption{Overview of representative cross-modality applications.}
\label{tab:cross_modal}
\resizebox{0.95\textwidth}{!}
{%
\begin{tabular}{llm{0.8cm}m{1.4cm}m{5.8cm}} 
\toprule
\textbf{Method} & \textbf{Tasks} & \textbf{Organ} & \textbf{Modality} & \textbf{Pre-trained CLIP} \\
\midrule
FlexR \citep{keicher2022flexr} &  Report Generation & Chest & X-ray & Fine-tuned CLIP~\citep{radford2021learning} \\
\midrule

MCGN \citep{wang2022medical} & Report Generation & Chest & X-ray & CLIP~\citep{radford2021learning} \\
\midrule

\textcolor{highlight}{TUMSyn}~\citep{wang2024towards} & \textcolor{highlight}{Image Generation} & \textcolor{highlight}{Brain} & \textcolor{highlight}{MRI} & \textcolor{highlight}{CLIP}~\citep{radford2021learning} \\
\midrule

\textcolor{highlight}{ContextMRI}~\citep{chung2025contextmri} & \textcolor{highlight}{Image Generation} & \textcolor{highlight}{Brain, Knee} & \textcolor{highlight}{MRI} & \textcolor{highlight}{CLIP}~\citep{radford2021learning} \\
\midrule

X-TRA \citep{van2023XTRA} & Image-text Retrieval & Chest & X-ray & Fine-tuned CLIP~\citep{radford2021learning}, \newline Fine-tuned PubMedCLIP \citep{eslami2021does} \\
\midrule

MONET \citep{kim2024transparent} & Image-text Retrieval  & Skin & Dermoscopy  & CLIP~\citep{radford2021learning} \\
\midrule

VQA-adapter \citep{liu2023parameter} & MedVQA & General  &  General  & CLIP~\citep{radford2021learning} \\
\midrule

 \cite{eslami2021does} & MedVQA &  General &  General & PubMedCLIP~\citep{radford2021learning} \\
\midrule

\citet{van2023open} & MedVQA& General  & General  & CLIP~\citep{radford2021learning} \\
\bottomrule 
\end{tabular}
}
\end{table*}

Following \cite{liu2023clip}, \cite{zhang2023continual} extend this framework into continual learning by leveraging additional heads and text prompts to tackle new tasks. As previous studies~\citep{ozdemir2018learn, ozdemir2019extending, liu2022learning} have focused on developing novel loss functions as additional constraints, or memory modules to preserve patterns from the original data, ~\cite{zhang2023continual} leverage text prompts to represent the correlation between current task and previously learned tasks. The semantic correlation contained in text prompts enables the proposed model to filter and reserve task-specific information with superior performance.

\subsubsection{Others}
{Different from fully supervised segmentation, weakly supervised segmentation aims to utilize weak annotations, e.g., points~\citep{bearman2016s}, scribbles~\citep{lin2016scribblesup, vernaza2017learning}, bounding boxes~\citep{dai2015boxsup, khoreva2017simple}, and text or description for image-level labels~\citep{kolesnikov2016seed, ahn2018learning}, as supervision for training.}
~{Class activation map (CAM) is a commonly-used solution for this task~\citep{zhou2016learning}, which equipped the convolutional neural network (CNN) with the ability for attention localization and pseudo-label generation.} Yet CAM only focuses on the most distinctive parts of the object and often leads to low-quality pseudo labels due to boundary neglection,~{especially in medical cases where the boundary between target foreground and background is ambiguous~\citep{chen2022c}.}
Despite recent attempts to broaden CAM's coverage~\citep{lee2021railroad, han2022multi, li2022online, zhang2022transws}, this fundamental problem persists. Notably, ~\cite{zhang2023tpro} propose to integrate language prior knowledge into weakly supervised learning to provide reliable assistance in finding object structures. Specifically, they introduce a text-prompting-based weakly supervised segmentation method (TPRO) by employing a pre-trained MedCLIP~\citep{medpixclip} to convert semantic labels into class-level embedding. 
An additional BioBERT~\citep{lee2020biobert} is also adopted to extract detailed information about the label's corresponding text descriptions.
These additional text supervisions are then fused with the image features, effectively improving the quality of pseudo labels and thus providing superior performance compared to other CAM-based methods.

Considering keypoint localization, various methods~\citep{o2018attaining, payer2020coarse, wu2023multi} have been developed for challenges in the medical imaging domain. While these methods have demonstrated solid performance in many cases, they still struggle to handle complex localization environments, ~{i.e., with incomplete structures~\citep{lessmann2019iterative}.}
TCEIP~\citep{yang2023tceip} integrates the instructional text embedding of the target region into a regression network to guide the prediction of implant position. By leveraging CLIP, TCEIP is able to interpret and process instructions such as 'left', 'middle', and 'right' alongside visual data, ensuring more precise and context-aware results. Its performance has surpassed the capabilities of previous image-only methods, especially in challenging cases with multiple missing teeth or sparse teeth.

\subsection{Cross-modal tasks}
In addition to previously mentioned pure vision tasks, CLIP has also propelled the development of cross-modal tasks. Here cross-modal task refers to the interaction between image and text modalities. Representative studies are shown in Table~\ref{tab:cross_modal}.

\subsubsection{\textcolor{highlight}{Generation}}
\textcolor{highlight}{The application of CLIP in generation can be categorized into report generation and image generation.}

\textcolor{highlight}{\textbf{Report generation.}} Given the time-intensive process of manually transcribing reports in clinical settings, there has been an increasing inclination toward automating the generation of medical reports~\citep{liu2019clinically, yu2023evaluating}. Since effective generation of medical reports needs the recognition of crucial findings, attributes, and inter-finding semantic relations, CLIP is inherently suitable for this task due to its alignment with human cognition and semantic awareness.
For example, ~\cite{wang2022medical} adopts CLIP's vision encoder to extract semantic-aware image representations from chest X-rays, which then interacts with learnable concept embeddings and hence benefits the performance of report generation. 
~\cite{keicher2022flexr} fully leverages the strengths of CLIP by reformulating the report generation task as a multi-label classification task, with labels indicating the presence or absence of specific findings. They compile all possible combinations of clinical findings and corresponding locations in the training set to form a prompt set. They utilize CLIP's zero-shot capability to calculate the likelihood of each prompt appearing in the image.

\textcolor{highlight}{\textbf{Image generation.} Given that CLIP can bridge the gap between image and language, it is also intuitive to apply this technique to text-guided image generation. In this context, ~\citet{wang2024towards} make the first attempt to generate customized MRI sequences from routinely acquired scans based on text prompts. Their proposed method, Text-guided Universal MR Image Synthesis (TUMSyn), enables the generation of MRI sequences tailored to specific text-based instructions. These prompts include subject information such as age and gender, along with  MRI parameters like scanning field strength, scanner type, voxel size, inversion time, etc. Their pre-trained CLIP text encoder is responsible for converting these text prompts into feature embeddings that can be interpreted by the LIFF model~\citep{chen2021learning}, achieving superior generation performance across several evaluation datasets. Likewise,~\citet{chung2025contextmri} address a specific inverse problem, compressed sensing MRI. They suggest that conditioning diffusion models on such metadata could significantly enhance the performance of this application.
}
\begin{figure}[tbp]
    \centering
    \includegraphics[width=0.45\textwidth]{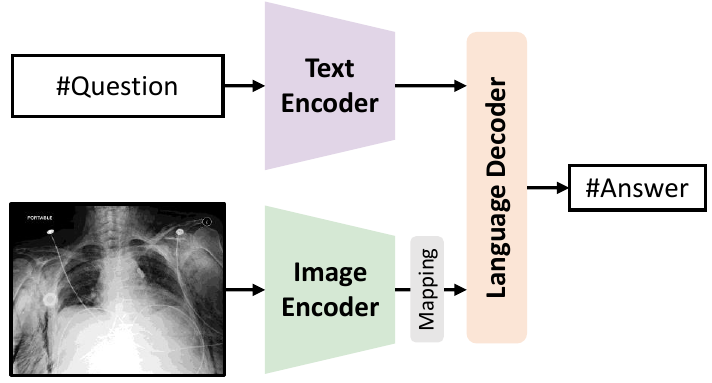}
    \caption{Illustration of CLIP-driven methods for open-ended MedVQA.}
    \label{fig:clip-VQA-tasks}
\end{figure}
\subsubsection{Medical visual question answering}
Medical visual question answering (MedVQA) is a task that demands an in-depth understanding of both text-based questions and visual content of medical images~\citep{lin2023medical}. It has drawn attention from the community as it would lead to more efficient and accurate diagnoses and treatment decisions. 
Since CLIP has long been favored by its ability to align visual and text content, recent efforts have been made to apply CLIP in MedVQA. 

Previous efforts have incorporated CLIP into closed-ended MedVQA tasks. These studies~\citep{eslami2023pubmedclip,liu2023parameter} usually only integrate CLIP's image encoder into the original MedVQA framework, aiming to enhance image representation with semantic understanding. However, they tend to overlook the comprehensive utilization of image-text alignment. Moreover, the closed-ended MedVQA typically offers all potential answer options for each question, which essentially transforms the task into a classification problem. Consequently, the practical utility of these methods appears constrained due to these limitations.

Conversely, the open-ended MedVQA has shown promise due to the development of CLIP. It does not pre-define options for each question, expanding its applicability to various scenarios and necessitating a heightened capacity for image-text comprehension~\citep{lin2023medical}. Hence, ~\cite{zhang2023pmc} leveraged CLIP's image encoder and text encoder for question and image understanding, respectively, with a subsequent language decoder for answer generation. We illustrate the CLIP-driven open-ended MedVQA in Fig.~\ref{fig:clip-VQA-tasks}. To mitigate the domain gap between CLIP's pre-training dataset and the current MedVQA dataset, a mapping layer is commonly employed. The question embedding and the transformed image embedding are concatenated and directly input into a language decoder, which may take the form of a multi-layer transformer or a language model, facilitating the generation of answers.

\subsubsection{Image-text retrieval}
\begin{figure}[tbp]
    \centering
    \includegraphics[width=0.48\textwidth]{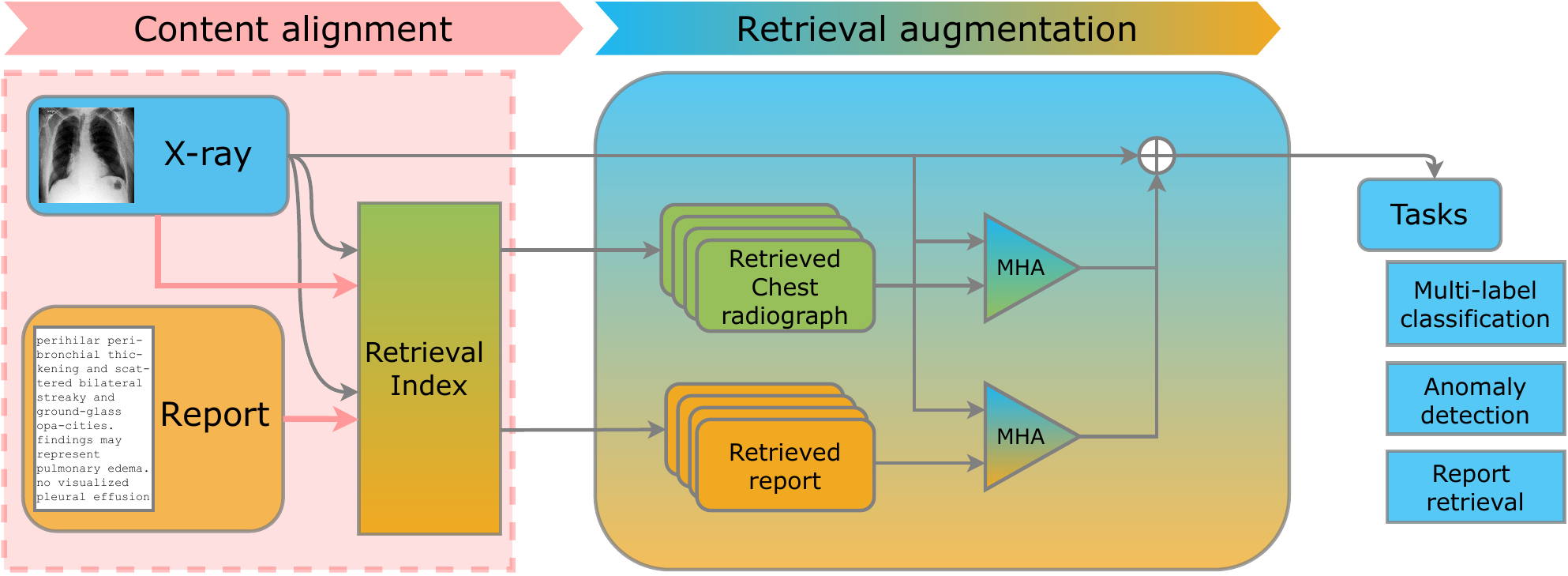}
    \caption{Architecture overview of X-TRA (from~\citet{van2023XTRA}).}
    \label{fig:clip-retrieval}
\end{figure}

 \begin{figure*}[t]
        \centering
        \includegraphics[width=0.95\linewidth]{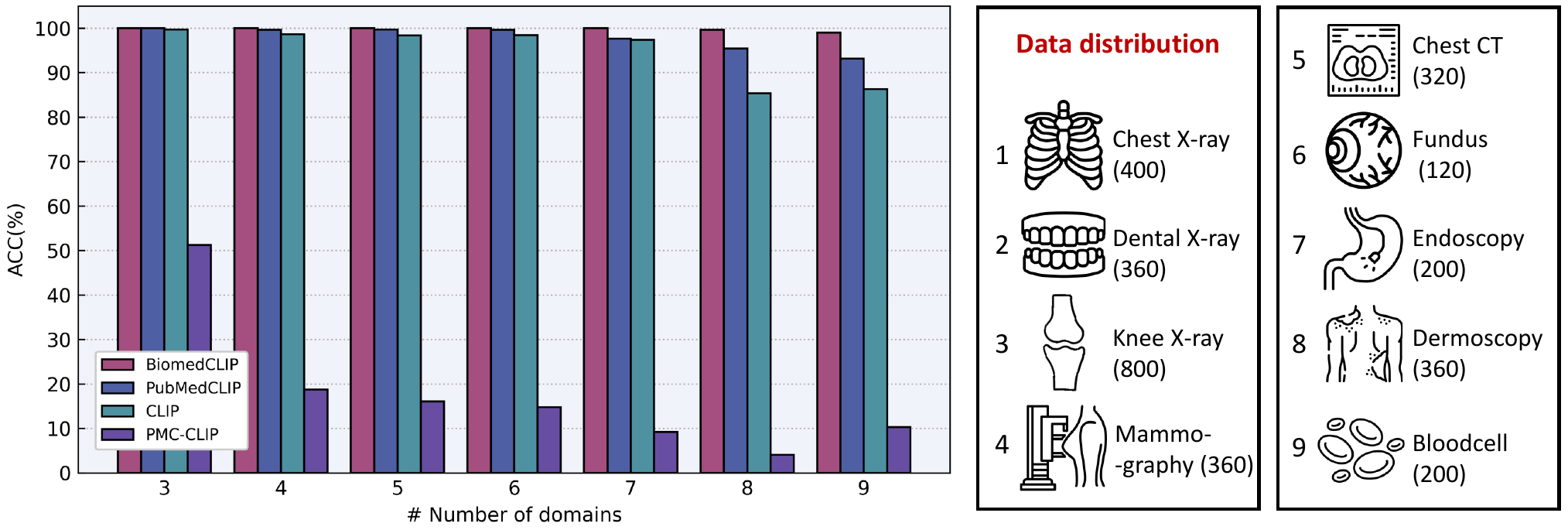}
        \caption{Quantitative analysis of different CLIP models on the task of domain identification. Pre-trained CLIP models were asked to discriminate the domain of each input image via zero-shot inference, which can, to some extent, reflect their understanding of general biomedical knowledge.}
        \label{fig:clip_quan_cmp}
    \end{figure*}

\begin{table*}[htbp]
  \centering
  \renewcommand{\thetable}{6}
  \caption{Comparison of Multi-modality Large Language Model (MLLM) and CLIP. Five aspects, including implementation efficiency, adaptability to different tasks, and user-friendliness were investigated.}
    \begin{tabular}{ccccccc}
    \toprule
          & Lightweight & Single-modality  & Multi-modality& Feature embedding & Interactive  \\
    \midrule
    MLLM  & \XSolidBrush & \XSolidBrush  &  \CheckmarkBold&  \XSolidBrush &  \CheckmarkBold  \\
    CLIP  & \CheckmarkBold& \CheckmarkBold  & \CheckmarkBold & \CheckmarkBold & \XSolidBrush  \\
    \bottomrule
    \end{tabular}
  \label{tab:clip_mllm}%
\end{table*}%
Retrieval augmentation~\citep{komeili2022internet}, which involves supplementing data by retrieving relevant information, allows utilization of up-to-time information from a trusted knowledge source, essentially providing a non-parametric memory expansion~\citep{ramos2023smallcap}. This approach has gained attention for its versatility, especially in the area of retrieval-augmented large language model~\citep{zhao2024chatcad+,asai2023retrieval}. However, existing retrieval methods often focus on global image features~\citep{clef2023overview}, which can lead to sub-optimal results in medical imaging. Unlike global features that may resemble across patients, 
subtle image details have effects on disease diagnosis and are of significance.

To address the domain shift between medical and natural images, \citet{van2023XTRA} propose a CLIP-based multi-modal retrieval framework. This method comprises two main parts as illustrated in Fig.~\ref{fig:clip-retrieval}. The first part involves finetuning the original CLIP model to construct the retrieval model. Given the visual similarity of medical images and the significance of small, localized markers as disease indicators, they propose a content classifier to implement supervised content-based alignment. The second part utilizes the output of the retriever in cross-modal retrieval augmentation, enhancing downstream tasks with multi-head attention (MHA).
When evaluating the performance of their retrieval method in comparison to previous approaches for disease classification and report retrieval, \cite{van2023XTRA} demonstrate substantial performance improvement, outperforming all existing retrieval methods by a significant margin. The observed performance difference underscores the potential of CLIP in constructing a robust retrieval method.

\subsection{Summary}
In this section, we demonstrate some representative CLIP-driven applications to show the performance improvements under CLIP assistance. While these studies focus on various tasks, they generally indicate the strength of a pre-trained CLIP model lying in its ability to \textbf{interpret and convey human knowledge}. As best illustrated in~\citet{pellegrini2023xplainer,zhang2023tpro,yang2023tceip}, where descriptive text prompts are fed to the CLIP, the experimental results showcase CLIP's adeptness in comprehending the semantics embedded within prompts and effectively conveying semantics to other components within the framework.
It implies that CLIP-driven applications can be adaptable to different groups of patients by simply modifying the specific content of the input prompt, which is beneficial for the diagnosis or prognosis of diseases having regional or age-related differences. For instance, diseases like sepsis often exhibit distinct progression patterns among different racial groups~\citep{khoshnevisan2021unifying, tintinalli2016tintinalli}, while the survival rate of Community-Acquired Pneumonia correlates with the patient's age~\citep{stupka2009community,ravioli2022age}. By adjusting the content within the descriptive prompt, the develop CLIP-driven application can seamlessly transition between various groups without requiring re-training or fine-tuning.

\section{Comparative analysis}

{
While we have investigated the advancement of two research lines in the above, this section aims to provide more practical guidance for the community by conducting a comparative analysis. We begin with a comparison between CLIP and other vision-language models to rationalize the utilization of CLIP in medical imaging, then we quantitatively evaluate the effectiveness of different CLIP models.}

\begin{figure*}[!tbp]
    \centering
    \includegraphics[width=0.95\textwidth]{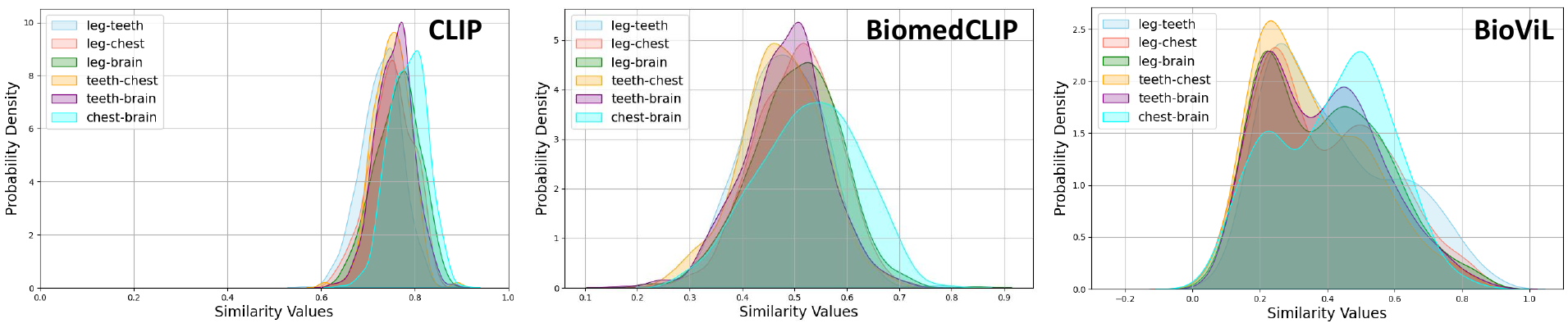}
    \caption{Comparison between non-domain-specific and domain-specific pre-training. CLIP (OpenAI) tends to present high inter-disease similarity, which is significantly alleviated in BiomedCLIP and BioViL, revealing the irrationality of adopting CLIP (OpenAI) in medical applications.}
    \label{fig:CLIP_cmp}
\end{figure*}

{
    \textbf{Strengths of CLIP over other vision-language models.} As mentioned in a concurrent study surveying medical foundation models~\citep{azad2023foundational}, a variety of vision-language foundation models have emerged since 2021, including Flamingo~\citep{alayrac2022flamingo}, BLIP-2~\citep{li2023blip}, and LLaVA~\citep{liu2024visual}. These non-CLIP foundation models typically leverage an image encoder and a pre-trained large language model, mapping the latent space of the image encoder to the language latent space via projection techniques such as gated xattn-dense layers (in Flamingo), Q-Former (in BLIP-2), or MLPs (in LLaVA). For simplicity, we categorize these vision-language models as Multi-modality Large Language Models (MLLMs). MLLMs have received significant attention from the community and have been specifically tailored for healthcare applications ~\citep{moor2023med,van2024large,li2024llava}. However, we argue that, at least currently, CLIP offers more benefits to the medical imaging domain than its competitors.  The comparison between MLLM and CLIP is demonstrated in Table~\ref{tab:clip_mllm}. } 

{
Firstly, due to the presence of large language models, MLLMs have an extremely high parameter count, making them difficult to deploy on edge computing devices. This hinders their widespread application in hospitals, especially in underdeveloped regions. In contrast, CLIP consists of only an image encoder and a text encoder, requiring fewer computational resources. Secondly, CLIP can adapt to a wider variety of tasks, offering greater flexibility. Since the image encoder and text encoder are independent components, CLIP can handle single-modality tasks or tasks involving multiple modalities. However, in MLLMs, the image encoder and language decoder are tightly connected, making it unsuitable for single-modality tasks and preventing it from providing high-quality feature embeddings for downstream tasks, thus limiting its applicability. On the other hand, we must acknowledge that MLLMs excel in interactivity, providing direct responses to instructions and offering a better user experience, a functionality that CLIP lacks.}

{
In summary, CLIP's lower computational resource requirements and broader range of application scenarios currently give it clear advantage over MLLMs in medical AI, despite its lack of direct user interaction capabilities.
    }

{
    \textbf{Comparison between different pre-trained CLIP models.} The quality of the selected pre-trained CLIP models is crucial for CLIP-driven applications, motivating us to conduct a quantitative analysis. Specifically, we evaluated the performance of various CLIP models on a dataset collected by~\citet{zhao2024chatcad+}, which includes images from nine different domains as shown in the right panel of Fig.~\ref{fig:clip_quan_cmp}. We adhered to the evaluation protocol proposed by~\citet{zhao2024chatcad+}. Four pre-trained models were selected for comparison: BiomedCLIP~\citep{zhang2023large}, PubMedCLIP~\citep{eslami2023pubmedclip}, CLIP~\citep{radford2021learning}, and PMC-CLIP~\citep{lin2023pmc}.
}

 {
During the evaluation, we progressively increased the number of domains in the order illustrated in Fig.~\ref{fig:clip_quan_cmp} and reported the accuracy (ACC) at each step, leading to several noteworthy observations:
\begin{itemize}
    \item BiomedCLIP~\citep{zhang2023large}, pre-trained on an in-house 15M-scale dataset, unsurprisingly achieved the best performance. It demonstrated impressive results, maintaining 100\% accuracy as the number of domains increased from 3 to 7. Even with all nine domains involved, it achieved an accuracy of 98.9\%, significantly outperforming the second place (93.11\% by PubMedCLIP). In contrast, the performance of the original CLIP model, pre-trained only using web-crawled data, unfortunately degraded to 86.3\% when dealing with images from nine domains. These findings illustrate that BiomedCLIP possesses a deeper understanding of professional medical knowledge compared to other candidates.
    \item While PMC-CLIP~\citep{lin2023pmc} has shown promising results when fine-tuned for downstream tasks~\citep{lin2023pmc,zhang2023pmc}, it performed poorly on domain identification, even worse than CLIP trained on natural images. This indicates its lack of zero-shot capability.
\end{itemize}
In summary, BiomedCLIP is generally the most recommended one for CLIP-driven applications, consistent with the conclusions in~\citet{van2024large}. Its ability to connect visual features to specific imaging domains highlights its potential for applications in tasks involving multiple organs or imaging modalities. The implementation of the aforementioned experiments can be found on our \href{https://github.com/zhaozh10/Awesome-CLIP-in-Medical-Imaging}{Github} project page.
}

\section{Discussions and future directions}\label{sec: discussion}

The aforementioned sections have delved into research studies that either leverage a refined CLIP pre-training paradigm or showcase CLIP-driven clinical applications within the medical imaging field. Despite significant strides, there still exist several challenges and open questions. In this section, we summarize key challenges and offer discussions on potential future directions.

\textbf{Inconsistency between pre-training and application.} 
Some readers might notice that the two sections -- refined CLIP pre-training and CLIP-driven application -- are currently uncoordinated. Ideally, refined CLIP pre-training is responsible for offering the domain-specific CLIP for CLIP-driven applications. 
Unfortunately, the CLIP-driven applications covered in this survey still primarily rely on OpenAI's pre-trained CLIP (trained on natural image-text datasets){, accounting for a high proportion of 76.67\%.} This would significantly limit their performance in clinical practice. 
In Fig.~\ref{fig:CLIP_cmp}, we select the top 20 to 30 most frequently occurring diseases for each organ, computing inter-organ similarity distributions of textual disease embeddings. Despite the inherent semantic differences among the selected diseases, the resulting similarity distribution reveals a challenge for CLIP in effectively discriminating them, as the distribution curves overlap significantly in the left panel of Fig.~\ref{fig:CLIP_cmp}. Notably, this issue is markedly mitigated in BiomedCLIP, underscoring the importance of domain-specific CLIP pre-training.
Simultaneously, BioViL, a model tailored for Chest X-ray analysis, demonstrates the best performance. This observation underscores the efficacy of specialized pre-trained CLIP model, emphasizing their ability to outperform generalized counterparts, especially in contexts where fine-grained discrimination among diseases is crucial.
Hence, we argue that future works focusing on CLIP-driven applications should adopt a pre-trained CLIP specific to their target organ.  
Even for research studies focusing on context optimization (in Section~\ref{sec: context_opt}), which aim to efficiently fine-tune a non-domain-specific CLIP to specific medical imaging scenarios, we still recommend the use of BiomedCLIP~\citep{zhang2023large} rather than OpenAI's general-purpose CLIP.

\textbf{Incomprehensive evaluation of refined pre-training.}
As previously illustrated in Section~\ref{sec: pre-train-summary}, studies focusing on refined CLIP pre-training commonly assess the quality of pre-training through various evaluation tasks. These evaluation tasks include those primarily aimed at evaluating vision encoders, such as CLS/ZSC/SEG/DET, and tasks that simultaneously assess image and text encoders, such as RET/VQA/PG. However, the issue lies in the fact that existing studies tend to favor vision-biased evaluation tasks, somewhat overlooking the evaluation of text encoders, which is evidently demonstrated in Table~\ref{tab:pre-training-paper}. 
The essence of CLIP lies in the alignment between images and texts. Only when both the vision encoder and the text encoder demonstrate high quality, they can then effectively function as foundational components in the domain-specific CLIP-driven applications. BioViL~\citep{boecking2022making} and BioViL-T~\citep{bannur2023learning} deserve recognition as they implement relatively comprehensive evaluations for their pre-trained vision and text encoder, and BioViL has been adopted in some CLIP-driven applications due to its robust performance(see Table~\ref{tab:classification}). 
For future work, we encourage researchers to conduct more comprehensive evaluations.
These evaluations could encompass their performance across tasks such as report generation (IU-Xray~\citep{pavlopoulos2019survey}), phrase grounding (MS-CXR~\citep{boecking2022making}), and VQA (EHRXQA~\citep{bae2023ehrxqa}).

{
\textbf{Challenges of volumetric imaging.}
While numerous studies have focused on X-rays and histopathology images, volumetric imaging modalities such as CT and MRI have received less attention despite their significant clinical utility. Several factors contribute to this disparity: (1) High computational demands for processing volumetric data. CT and MRI scans, being higher-dimensional data, often require GPUs with over 40GB of VRAM. For example, ~\citet{hamamci2024foundation} employed four A100 GPUs, each with 80GB of VRAM, for the pre-training of their proposed CT-CLIP model. (2) Limited availability of large-scale annotated volumetric datasets for training and evaluation. For instance, CT-RATE~\citep{hamamci2024foundation}, currently the largest public dataset with paired CT scans and reports, contains only 50,000 CT scans (Table~\ref{tab:dataset}).
(3) Lack of semantic correspondence. Volumetric imaging encompasses richer information compared to 2D modalities. However, this exacerbates the issue of multi-scale features (see Section~\ref{sec: global-local}), as it becomes more difficult for sentence-level features in diagnostic reports to adaptively find their visual correspondences in the data volume, which will adversely impact the effectiveness of image-text alignment.
}

{
While the challenge of computational resources is difficult to address through technical solutions, we would like to suggest potential directions for addressing the remaining two challenges.
For the lack of large-scale volumetric datasets, synthetic data has shown promising results in the medical imaging domain and could be a viable solution. Synthetic data generation, as highlighted by~\citet{gao2023synthetic,shen2023image,hu2023label}, can effectively mitigate the imbalance in disease category distributions~\citep{hu2023label,holste2024towards}, thereby enhancing the generalizability of deep learning models. Currently, several studies have been devoted to the synthesis of volumetric imaging data~\citep{hamamci2023generatect,dayarathna2023deep,xu2024medsyn,ou2024synthesizing}. The advancements in this field are expected to facilitate the application of CLIP.
Addressing the challenge of semantic correspondence, eye-tracking data presents an innovative solution. Studies such as ~\citet{wang2022follow,zhao2024mining,kong2024gaze,kumar2024improving} demonstrate how eye-tracking can reveal semantic-correlated visual and text features. During medical imaging interpretation, radiologists explore images while simultaneously dictating findings~\citep{bigolin2022reflacx}. This concurrent process enables straightforward association of each sentence in the radiologist’s dictated report with its corresponding visual regions using gaze data, thereby facilitating effective multi-scale alignment for volumetric imaging.
     } 

\textbf{Limited scope of refined CLIP pre-training.}
Presently, domain-specific CLIP models are tailored specifically for Chest X-rays within medical imaging, leaving other prevalent image types like mammography, knee MRI, and histology without adequate research. This limitation is primarily attributed to the scarcity of publicly available medical datasets. Previously, MIMIC-CXR stood as the predominant large-scale dataset for image-text alignment in medical imaging. However, with the recent release of FFA-IR (in 2021) and two additional histology datasets (in 2023), there is a pressing need for further advancements in CLIP pre-training that prioritize these two domains rather than solely focusing on chest X-rays. These two domains also have their specific challenges, which make them different from chest X-rays. The FFA-IR dataset is featured by multi-view diagnosis. A fundus fluorescein angiography (FFA) examination may include tens or even more images to comprehensively assess the status of the eye's vascular system, which is much larger than that of Chest X-ray (only 1 or 2 views). At the same time, histology images are characterized by giga-pixel resolution and are usually processed at the patch level, which encourages the investigation of patch-level alignment and slide-level alignment. 
We expect that future works could develop more sophisticated CLIP-style pre-training methods to address these issues on domains beyond Chest X-ray.

{
    \textbf{Debiasing in CLIP Models.} The rapid advancement of CLIP-based models has brought to light their impressive capabilities across various tasks. However, it has also revealed inherent biases that can affect the fairness and accuracy of these models.
Recent research has shown that CLIP-based models can exhibit significant biases. These biases often arise from the data on which the models are trained and the design of the models themselves. For instance, studies by ~\citet{luo2024fairclip} and ~\citet{zhang2022contrastive_nips} have highlighted how CLIP models can reflect and even amplify societal biases present in the training data. These biases can manifest in various ways, such as:
\begin{itemize}
    \item Gender and racial biases: CLIP models may exhibit preferential treatment or stereotyping based on gender, race, or other demographic factors, affecting the fairness of the model's predictions.
    \item Domain-Specific biases: In medical imaging, biases can arise from imbalances in the representation of different diseases or groups, leading to skewed performance.
\end{itemize}
In the context of medical imaging, bias in CLIP models can have serious consequences, including misdiagnosis or unequal quality of care. Ensuring that CLIP models are fair and unbiased is critical for their deployment in healthcare applications. Addressing these biases not only improves the accuracy and reliability of the models but also helps in building trust and ensuring equitable healthcare delivery.
Future research should focus on refining debiasing techniques and applying them to medical imaging domains. This includes exploring how bias mitigation methods can be tailored to address specific challenges and requirements of medical applications.
    }

{
    \textbf{Enhancing adversarial robustness of CLIP.} CLIP’s notable zero-shot learning capabilities make it a powerful tool across various domains. However, recent research indicates that it is vulnerable to adversarial perturbations, which can significantly undermine its performance~\citep{mao2023understanding,wang2024pre,schlarmann2024robustclip}.
For example,~\citet{thota2024demonstration} demonstrated that PLIP~\citep{huang2023visual}, a foundation model pre-trained on histology images, is susceptible to adversarial noise. This vulnerability led to severe misclassifications, emphasizing that even models trained on domain-specific data are not immune to adversarial attacks. Similarly,~\citet{wang2024pre} highlighted that adversarial perturbations can degrade CLIP's performance across various tasks thereby posing significant challenges for its deployment in critical applications.
Enhancing the adversarial robustness of CLIP is not only essential for maintaining its accuracy and reliability but also for ensuring its safe and effective application in sensitive areas like healthcare. Addressing these challenges promises to advance CLIP’s capabilities and extend its applicability in critical domains where robustness is paramount.
    }
\begin{figure}[t!]
    \centering
    \includegraphics[width=0.48\textwidth]{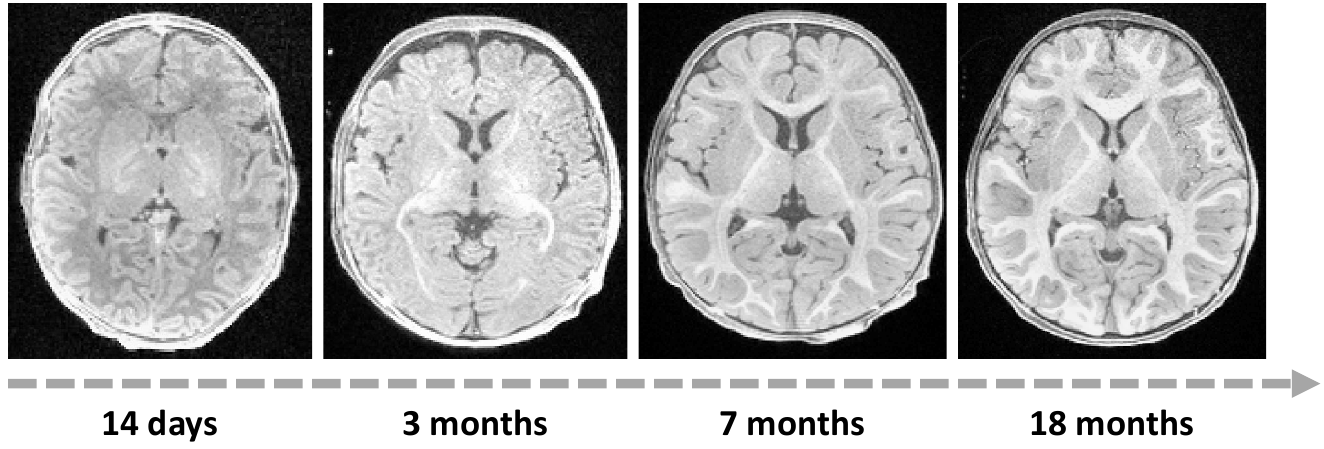}
    \caption{{Illustration of changes in both brain morphology and intensity contrast throughout infancy.}
    }
    \label{brain_volume_age}
\end{figure}

\textbf{Exploring the potential of metadata.} The potential of metadata has been largely underexplored. This type of data typically includes a range of patient attributes, many of which may exhibit a strong correlation with visual morphology. For example, 
{brain development is dynamic along with age \citep{stiles2010basics,wang2011automatic,li2014mapping,li2019computational}, especially during infancy.}
Fig.~\ref{brain_volume_age} illustrates the varied morphology and tissue contrast observed throughout infancy, highlighting the potential importance of age information in brain-related tasks. {Thus, integrating the metadata into the prompt may enhance the performance of deep learning models for brain development study.}
Unlike previous methods that encoded metadata directly using a multi-layer perceptron~\citep{cetin2023attri,zhao2021diagnose}, CLIP can offer a more semantically rich approach to text embeddings due to large-scale pre-training. This suggests a promising avenue for future research and exploration, and multiple studies have been conducted in the areas of AD diagnosis and brain MRI segmentation~\citep{teng2024knowledge} and Alzheimer's Disease (AD) diagnosis~\citep{liu2024progressive}.

\textbf{Incorporation of high-order correlations.} Existing CLIP-style pre-training methods in the medical imaging domain still predominantly adhere to orthogonal alignment between images and texts, lacking explicit consideration for inter-sample correlation. This conventional practice involves orthogonal alignment of each image with its corresponding ground-truth report. As elucidated in Section~\ref{sec: data-efficient}, this approach may result in performance degradation due to substantial semantic overlap among medical samples.
Although attempts have been made to mitigate this issue through inter-report semantic similarity, their success has primarily relied on handcrafted rules and low-order inter-report correlations. 
Consequently, the integration of high-order correlations emerges as a promising solution.

The effectiveness of high-order correlations has been well-established in tasks involving multiple information sources or those requiring interpretations of complex relationships, including brain network analysis ~\citep{ohiorhenuan2010sparse,chen2016high,zhang2017hybrid,owen2021high,liu2023learning}, multi-label classification~\citep{zhang2014multilabel,nazmi2020evolving,si2023multi}, and multi-view clustering~\citep{li2022high}. Likewise, medical image-text pre-training involves two kinds of information (i.e., image and text), and their semantic correlations need to be further explored. Hence, we expect that future studies will devote increased attention to comprehending the intricate semantic correlations between medical image-text samples,
addressing the challenge of orthogonal image-text alignment via the methodology of high-order correlation.

\textbf{Beyond image-text alignment.}
The philosophy of CLIP revolves around achieving alignment between different modalities, specifically images and text. Alignment, in this context, refers to the model's ability to understand and establish meaningful connections between visual and textual content. By comprehending the intrinsic connections between visual and textual information, CLIP can perform exceptionally well in various cross-modal applications, reflecting a broader trend.
Extending the alignment philosophy of CLIP to other multimodal medical imaging can be a promising direction. Medical imaging often involves various modalities like X-rays, MRI, and CT scans, each providing unique insights into different aspects of the same patient's condition. Analogous to CLIP's methodology, aligning these diverse imaging modalities within a unified embedding space could potentially revolutionize medical data analysis, presenting a progressive direction for medical research and diagnostics. This kind of technique could also be used for aligning same-modality images from different subjects as widely investigated by various image registration techniques before~\citep{jia2012iterative,fan2018adversarial}.

\section{Conclusion}\label{sec: conclusion}

In conclusion, we have presented the first review of the CLIP in medical imaging. Starting by introducing the foundational concepts that underpin CLIP's success, we then delve into an extensive literature review from two aspects: refined CLIP pre-training methods and diverse CLIP-driven applications.
For refined CLIP pre-training methods, our survey offers a structured taxonomy based on the unique challenges that CLIP pre-training owns in the medical imaging domain, aiming to chart a clear pathway for researchers to advance this field progressively.
In exploring diverse CLIP-driven applications, we compare CLIP-related approaches against those solely vision-driven methods, emphasizing the added value that pre-trained CLIP models could bring. Notably, through thoughtful design, they could serve as valuable supplementary supervision signals, significantly enhancing the performance across various tasks.
Beyond simply reviewing existing studies in these two sections, we also discuss common issues, laying the groundwork for future directions. By illuminating the potential and challenges of employing CLIP in medical imaging, we aim to push the field forward, encouraging innovation and paving the way for human-aligned medical AI.



\section*{Acknowledgments}
This work was supported in part by National Natural Science Foundation of China (grant numbers U23A20295, 82441023, 62131015, 82394432), the China Ministry of Science and Technology (S20240085, STI2030-Major Projects-2022ZD0209000, STI2030-Major Projects-2022ZD0213100), Shanghai Municipal Central Guided Local Science and Technology Development Fund (No. YDZX20233100001001), and HPC Platform of ShanghaiTech University.

\bibliographystyle{model2-names.bst}\biboptions{authoryear}
\bibliography{refs}



\end{document}